\documentclass[times,final,authoryear]{./elsarticle}

\usepackage{framed,multirow}

\usepackage{amssymb}
\usepackage{latexsym}
\usepackage{graphicx}
 \usepackage{array}
 \usepackage[table]{xcolor}
 \usepackage{gensymb}
 \usepackage{mathtools}
 \usepackage{adjustbox}
 \usepackage{float}
\usepackage{url}
\usepackage{xcolor}
\definecolor{newcolor}{rgb}{.8,.349,.1}

\journal{Pattern Recognition Letters}

\begin{document}

\thispagestyle{empty}

\begin{frontmatter}

\title{Facial Descriptors for Human Interaction Recognition In Still Images}

\author[a]{Gokhan {Tanisik}}
\author[a]{Cemil {Zalluhoglu}}
\author[a]{Nazli {Ikizler-Cinbis}\corref{cor1}} 
\cortext[cor1]{Corresponding author: 
  Tel.: +90-312-297-7500;  
  fax: +90-312-297-7502;}
\ead{nazli@cs.hacettepe.edu.tr}


\address[a]{Department of Computer Engineering, Hacettepe University, 06800 Ankara, Turkey}


\begin{abstract}

This paper presents a novel approach in a rarely studied area of computer vision: Human interaction recognition in still images. We explore whether the facial regions and their spatial configurations contribute to the recognition of interactions. In this respect, our method involves extraction of several visual features from the facial regions, as well as incorporation of scene characteristics and deep features to the recognition. Extracted multiple features are utilized within a discriminative learning framework for recognizing interactions between people. Our designed facial descriptors are based on the observation that relative positions, size and locations of the faces are likely to be important for characterizing human interactions. Since there is no available dataset in this relatively new domain, a comprehensive new dataset which includes several images of human interactions is collected. Our experimental results show that faces and scene characteristics contain important information to recognize interactions between people.

\end{abstract}


\end{frontmatter}



\section{Introduction}
\label{sec:intro}

In the last decade, human action recognition has been a very active research area in computer vision due to its various potential applications. A large body of work is dedicated to recognizing singleton activities in videos, whereas some recent work focus on recognizing singleton actions within still images. Human interaction recognition, which constitutes up a significant subset of multi-person activities, is a relatively less studied area. Especially for still images, the prior work is almost non-existent. In this paper, we address this problem of multi-person interaction recognition in images.  

When recognizing interactions in still images, the problem gets more complex and harder to solve, due to the explicit need to discriminate foreground from background clutter without the motion information. In videos, motion is shown to be a great clue for identifying the type of the interactions (\cite{wang2013,betterMotion}). Without motion, the foremost cue becomes the appearance. In this paper, we explore how we can extract and leverage multiple forms of appearance information for interaction recognition in images. 

\begin{figure*}
\begin{center}
   \includegraphics[width=0.15\linewidth]{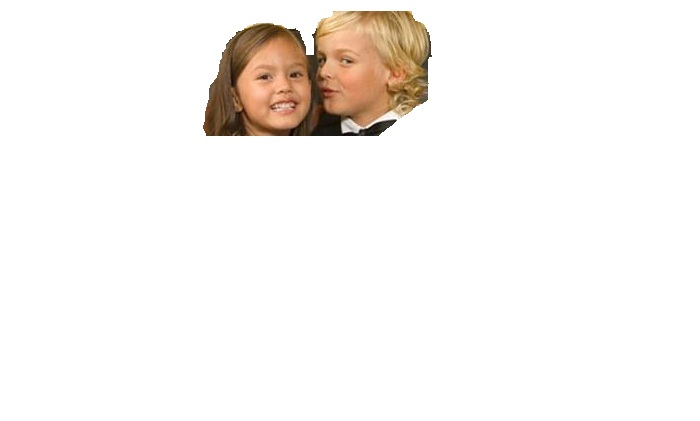}
   \includegraphics[width=0.15\linewidth]{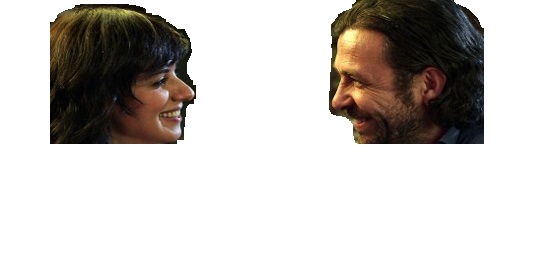}
   \includegraphics[width=0.15\linewidth]{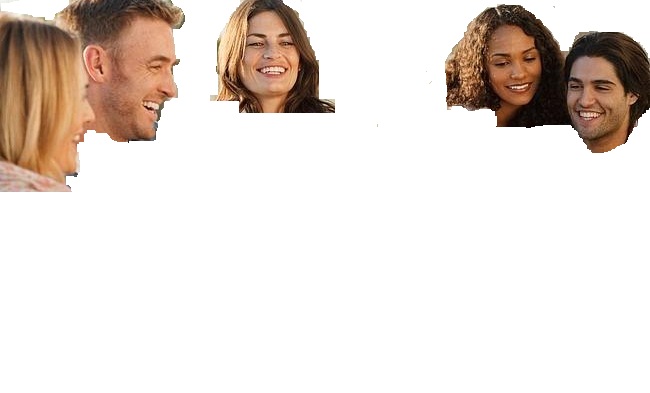}
   \includegraphics[width=0.15\linewidth]{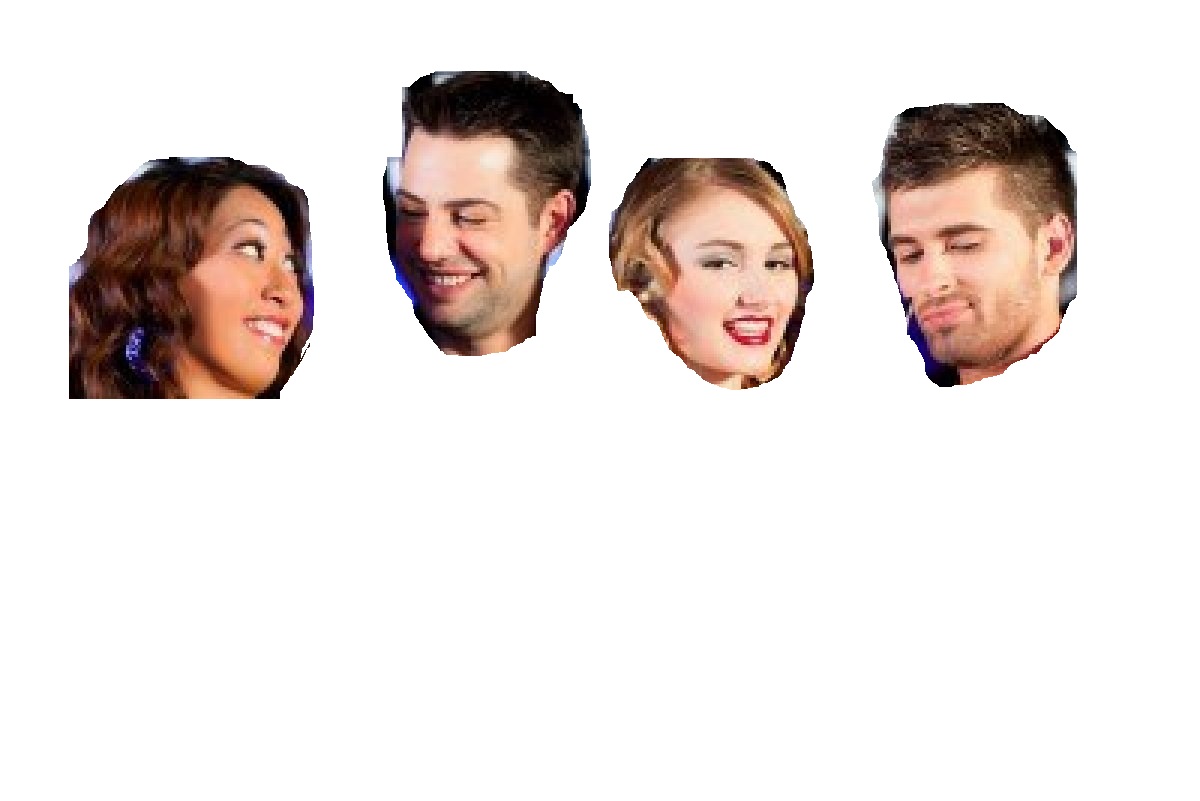}
   \includegraphics[width=0.15\linewidth]{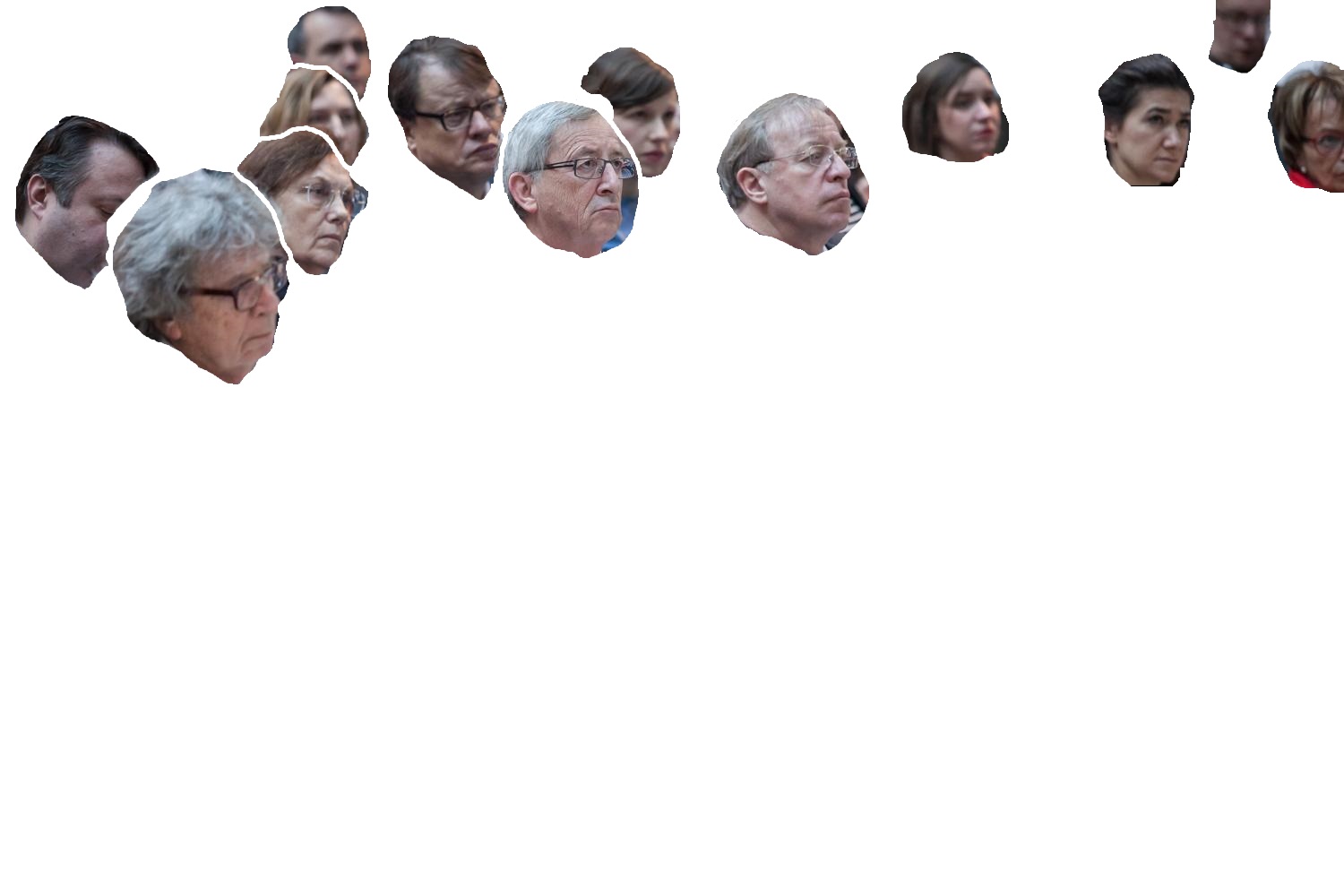}
   \includegraphics[width=0.15\linewidth]{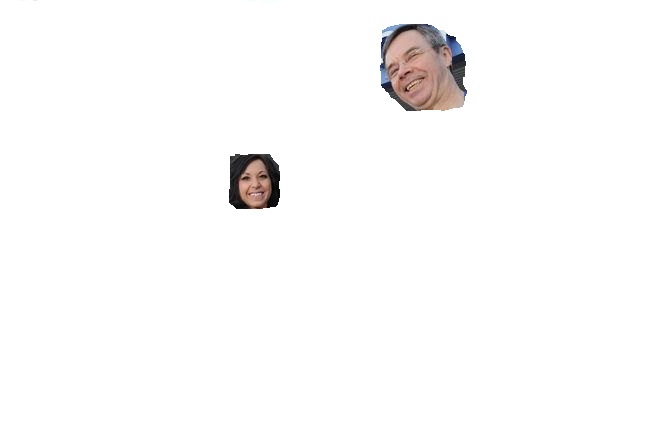}
\end{center}
\begin{center}
   \includegraphics[width=0.15\linewidth,height=1.5cm]{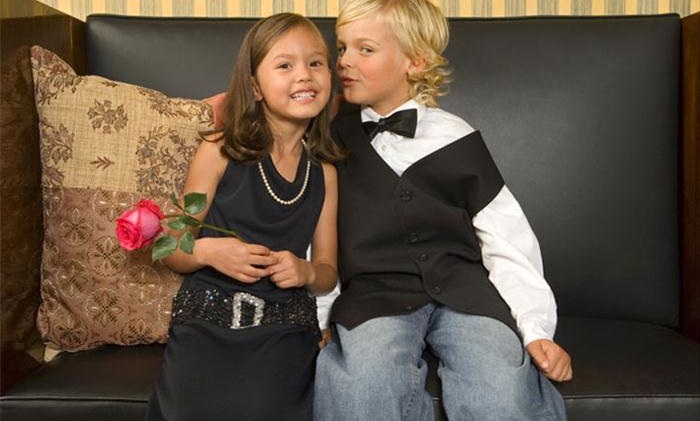}
   \includegraphics[width=0.15\linewidth,height=1.5cm]{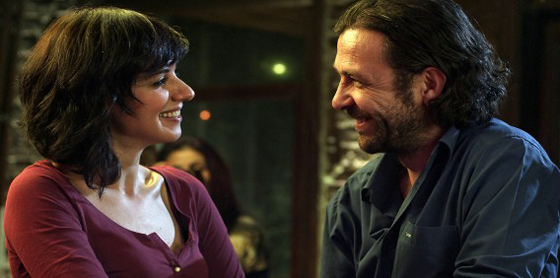}
   \includegraphics[width=0.15\linewidth,height=1.5cm]{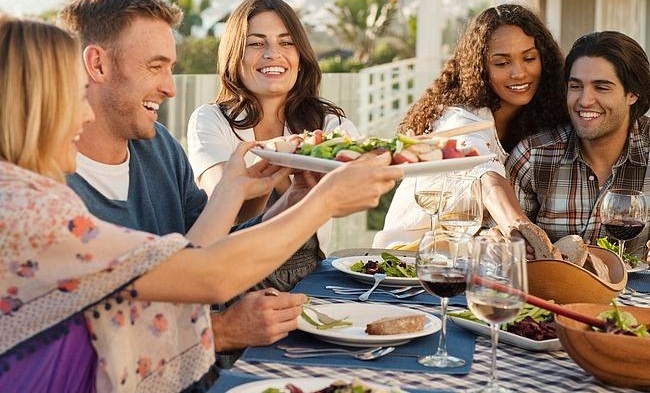}
   \includegraphics[width=0.15\linewidth,height=1.5cm]{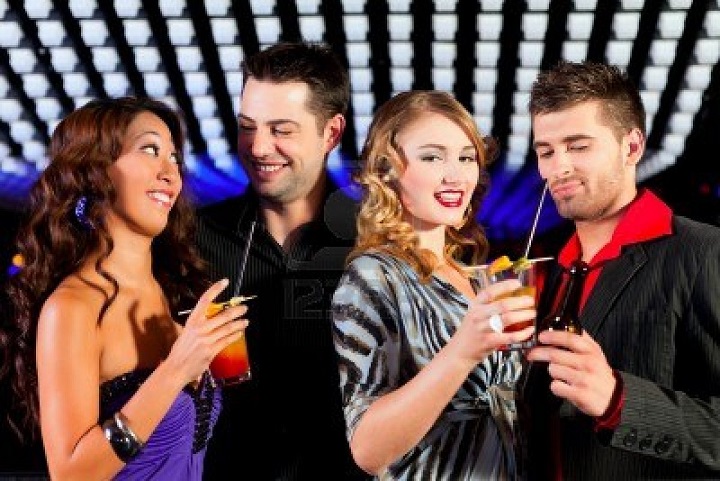}
\includegraphics[width=0.15\linewidth,height=1.5cm]{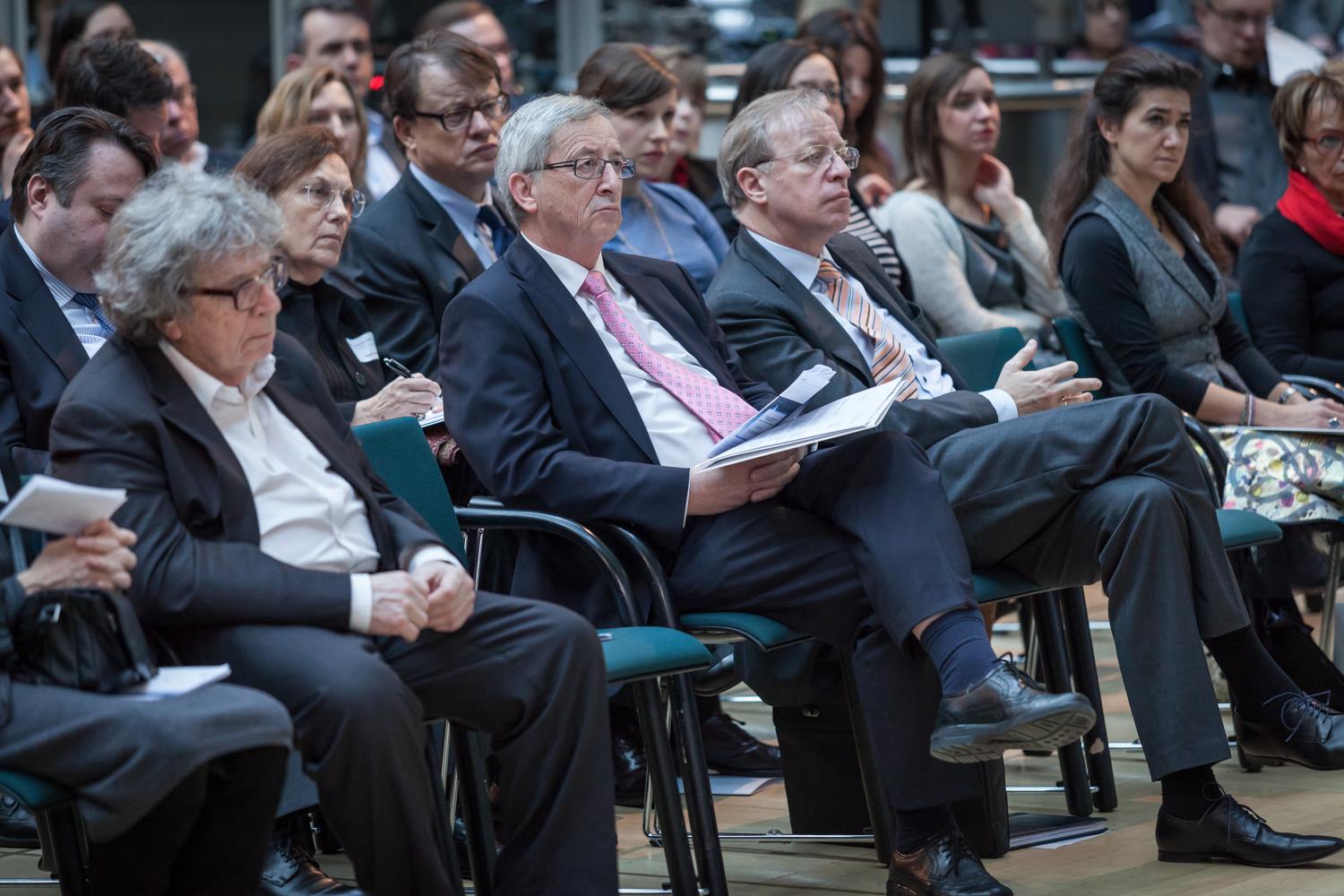}
   \includegraphics[width=0.15\linewidth,height=1.5cm]{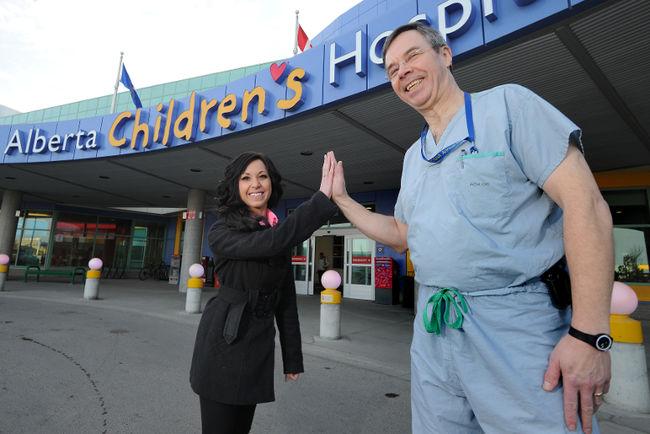}
   
\end{center}
\vspace{-0.5cm}
   \caption{Faces in interactions. The original images are shown in bottom row and the facial regions are shown on top. In this paper, we explore whether we can predict the type of human interactions in an image based on descriptors extracted from faces and their spatial layout. As it can be seen from this figure, without using any context or scene information, recognizing interactions by only face information can be quite a difficult task even for humans. Our results show that, using facial descriptors together with global scene descriptors yield promising results for human interaction recognition in still images.}	
\label{fig:interactionsWithOnlyFaces}
\end{figure*}

In this context, we propose several novel visual features that captures the intrinsic layout and orientation of face regions. Faces tend to play a great role in characterizing human interactions. People look at each other when they are talking, faces come together when people are kissing, and more. Figure \ref{fig:interactionsWithOnlyFaces} includes some examples. ~\cite{6247805} use faces in video sequences to describe interactions in a day-long first-person(egocentric) video of a social event. Based on inspiration from this work, we explore whether facial features can also be helpful in discriminating multi-person interactions in still images. 

Another reason to explore facial features is that, face detection technology is considerably advanced and is able to locate a great deal of faces in images, especially those that are not too small or significantly occluded. Our designed descriptors are based on the observation that relative positions, size and locations of the faces are likely to be important for characterizing human interactions. To extract these descriptors, we first use a face detector. We also estimate the orientations of the faces if possible using the face detector of~\cite{FaceDetection1}. In this way, we estimate the size, spatial location and the orientation of the face in a [-90\degree,90\degree] range with 15\degree resolution. We then make use of these features to propose image-level facial interaction descriptors. For recognition, we combine these multi-person facial descriptors with standard scene descriptors extracted globally from the images. In this context, we also investigate the effect of the state-of-the-art deep learning based features, aka, Convolutional Neural Network (CNN) features to this new problem domain.  

Since there is no available dataset in this relatively new domain, a new and comprehensive dataset, which includes a total of ten human interaction classes, such as boxing, dining, kissing, partying, talking, is collected. This dataset has also been enriched with the manual annotations of the ground truth face locations and orientations to facilitate further comparisons.  

Our contributions in this paper are two-fold: (1) We collect a new image dataset for human interaction categorization which includes multi-person interaction instances and (2) We present novel descriptors based on facial regions for human interaction recognition. 

Our experimental results show that, deep learning based features are effective in recognition of human-human interactions in images, and the proposed facial features that aim to encode the relative configurations of faces also provide useful information, especially when combined with global image features. In the rest of the paper, we first 

\section{Related Work}

There is a vast literature on human action/activity recognition (for a recent survey, see \cite{Aggarwal2011}), whereas human interaction recognition is a relatively less studied topic. There are a number of studies that propose models for interaction recognition in videos, and in general, two types of interactions are considered. These are human-object and human-human interactions. For human-object interaction recognition, \cite{Gupta2009} propose to use probabilistic models for simultaneous object and action recognition. For recognizing human-human interactions, \cite{ParkAggar2006} propose to simultaneously segment and track multiple body parts of interacting humans in videos. \cite{RyooAggar2009} builds their model on matching of local spatio-temporal features. \cite{JimenezIJCV2014} focus on a single type of interaction, i.e., looking at each other, and propose several methods for detecting this interaction effectively in videos. Recently,~\cite{HoaiCVPR14} utilize configuration detection of upper body detections for better interaction recognition in edited TV material. 

In image domain, human-object interactions are the focus of a number of studies, which handle the problem by extraction of distinctive feature groups~\cite{YaoFeiFeiCVPR2010a}, by bag-of-features and part-based representations~\cite{Delaitre10} and by weakly supervised learning~\cite{prest2012}. Object-person interactions have also been explored in~\cite{desai2010,YaoFeiFeiCVPR2010b,Delaitre2011}.

One of the earliest works to recognize the human interactions in still images is the work of \cite{905274}. In their paper, four classes are defined: shaking hands, pointing at the opposite person, standing hand-in-hand and intermediate-transitional state between them, and K-nearest neighbor classifier is used to recognize the interactions. Recently,~\cite{YangBKR12} has focused on how people interact by investigating the proxemics between them. They claim that complex interactions can be modeled as a single representation and a joint model of body poses can be learned. \cite{socialRole} look into the problem of detecting social roles in videos in a weakly supervised setting via a CRF model. In our work, we approach the human-human interaction recognition problem by means of several descriptors that encode facial region configurations. 

Another study area that could be related to our work is event recognition in still images(~\cite{6248037},~\cite{bossard13},~\cite{Li.dynpooling.2013}). Event recognition research aims to recognize a certain scene or event in images or videos. Datasets in this field are different from ours. Event recognition datasets describe an event like Christmas, wedding, etc. In such images, the main focus is not the people, but visual elements for an event. In multi-person interaction recognition problem, we focus on the presence of people, and try to infer the interaction based on images of people. 

In this work, we are inspired from the recent work of \cite{6247805}, which uses face detection responses to recognize social human interactions in video sequences from a first person perspective camera. They propose to use Markov Random Field for frame based feature representations and a Hidden Conditional Random Field to represent sequence based features. In our work, we propose several simple features based on face regions for recognizing human-human interactions in the images. 


\section{Our Approach}
In this section, we describe the facial descriptors and the learning procedure that we have proposed for the purpose of interaction recognition. 

\subsection{Visual Features for Human Interaction Recognition}
Our approach begins with the detection of the faces. For this purpose, we first apply the recent algorithm of \cite{FaceDetection1}, since it outputs three essential information about the faces: ~(1) Orientation of the face in the range of [-90\degree, 90\degree] with a resolution of 15\degree, ~(2) location of the face in the image and ~(3) size of the face in pixels. The orientation of a face is defined as the angle of the face with respect to the imaginary axis that crosses from the midpoint of the chin and the forehead. For reducing the number of false negatives in face detection, we also employ the OpenCV implementation of \cite{Viola2004}, which only gives the location and size of the images, and whether they are frontal or profile. The face detections from these two approaches are combined in the following way: ~(1) If both of the detectors find a face in the same region, \cite{FaceDetection1}'s output is used, since it is shown to be more accurate and it outputs face orientation estimates as well as face locations. ~(2) While using Viola-Jones face detector, if only frontal face is detected, the orientation is assumed to be 0\degree. If a profile face is detected, then the orientation is assumed to be 90\degree. If both frontal and profile face detectors fire within the same region, it means that the orientation of the face is between [0\degree, +/-90\degree]. To quantize the angle, the intersection ratio is normalized to [0\degree-90\degree] interval. 

After detecting the faces, we extract several mid-level descriptors based on the facial regions. Below, we introduce each of these descriptors. 

\vspace{2mm}

\begin{figure}[h!]
	\begin{center}
		\begin{tabular}{ccc}
			\includegraphics[width=0.2\linewidth,height=1.3cm]{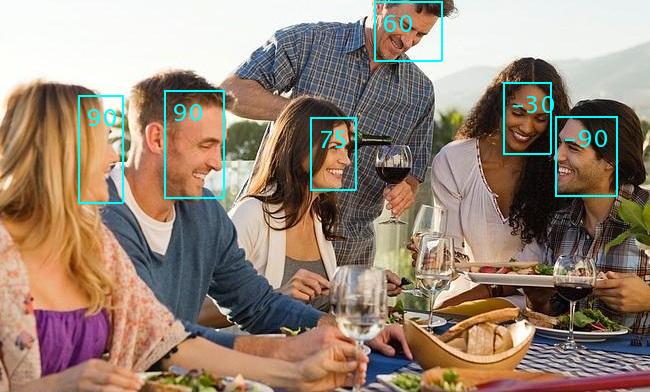}
\includegraphics[width=0.2\linewidth,height=1.3cm]{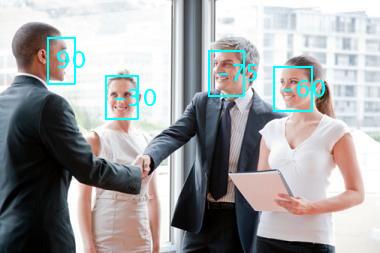}
			\includegraphics[width=0.2\linewidth,height=1.3cm]{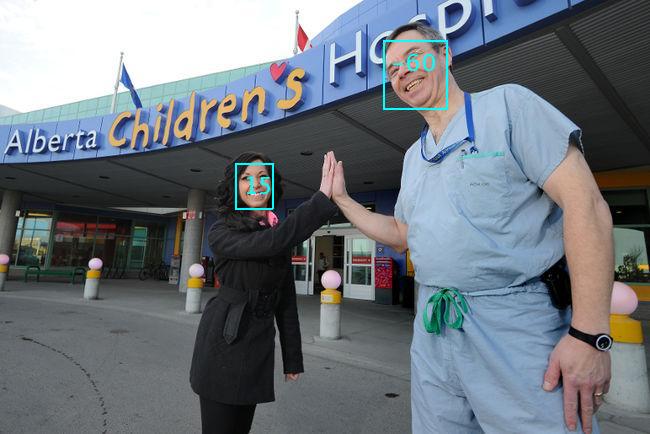}
		\includegraphics[width=0.2\linewidth,height=1.3cm]{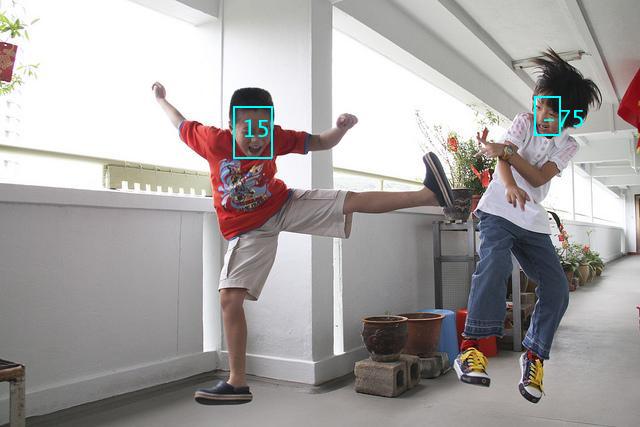}\\
			\includegraphics[width=0.2\linewidth,height=1.3cm]{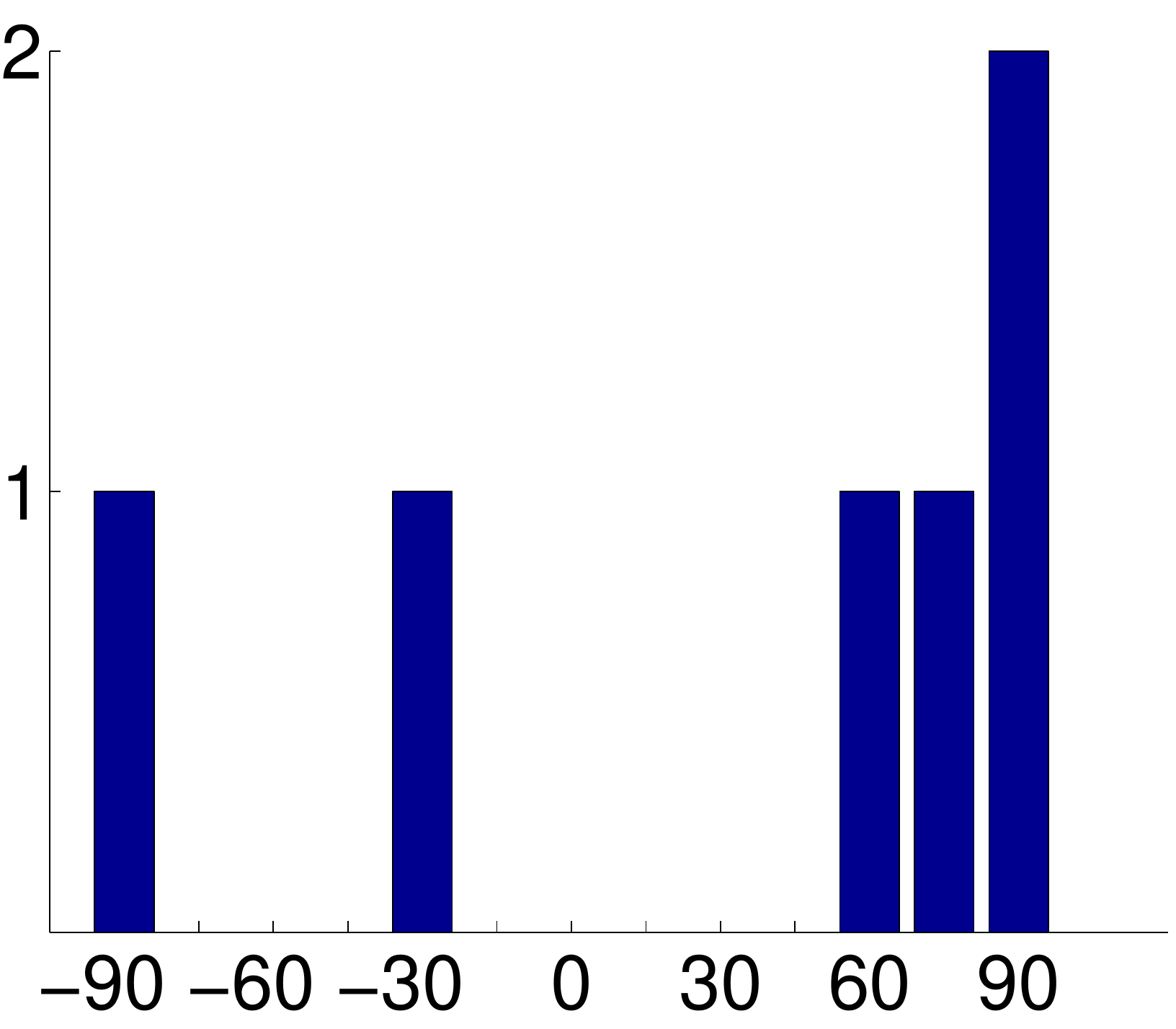}
			\includegraphics[width=0.2\linewidth,height=1.3cm]{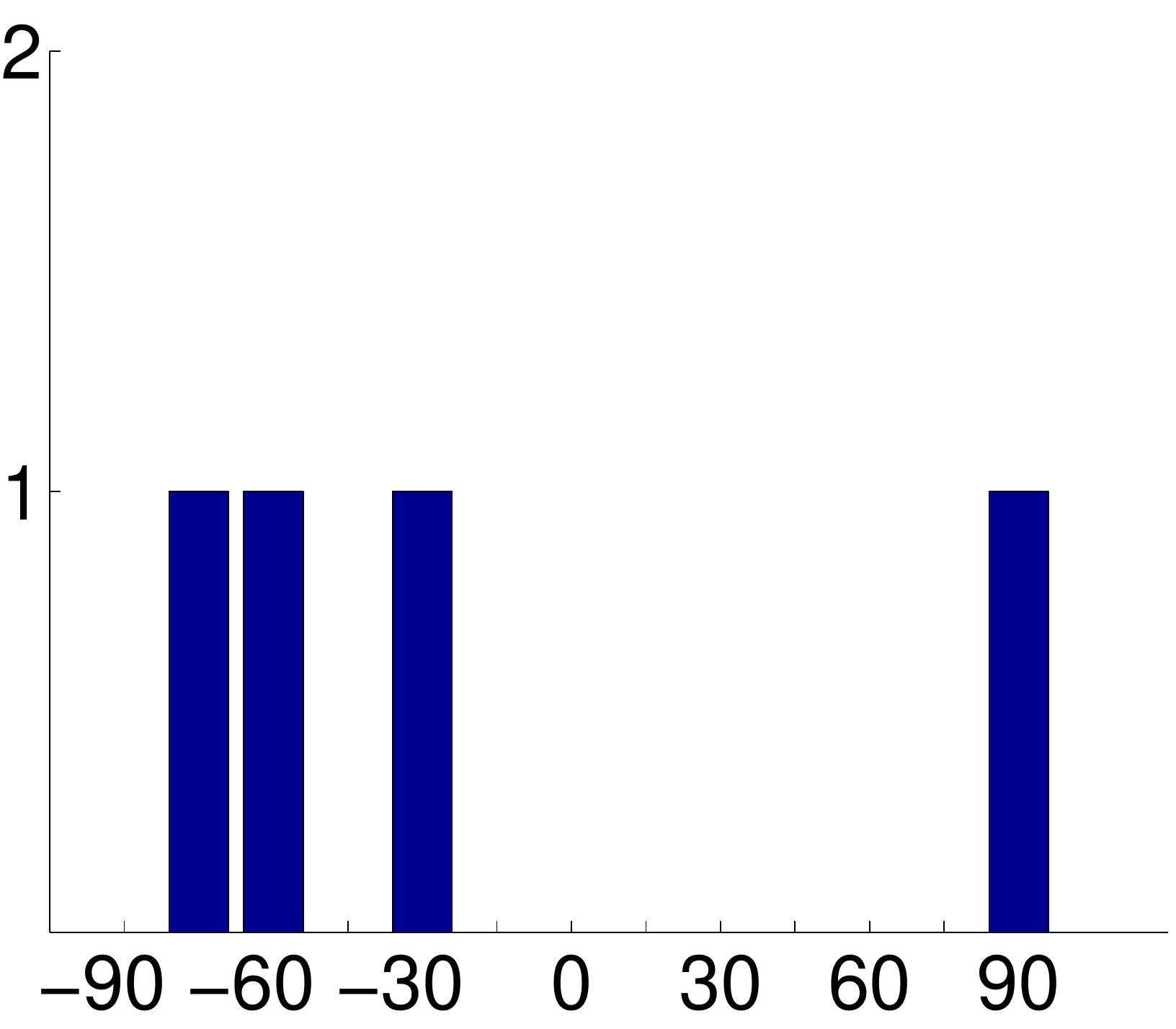}
			\includegraphics[width=0.2\linewidth,height=1.3cm]{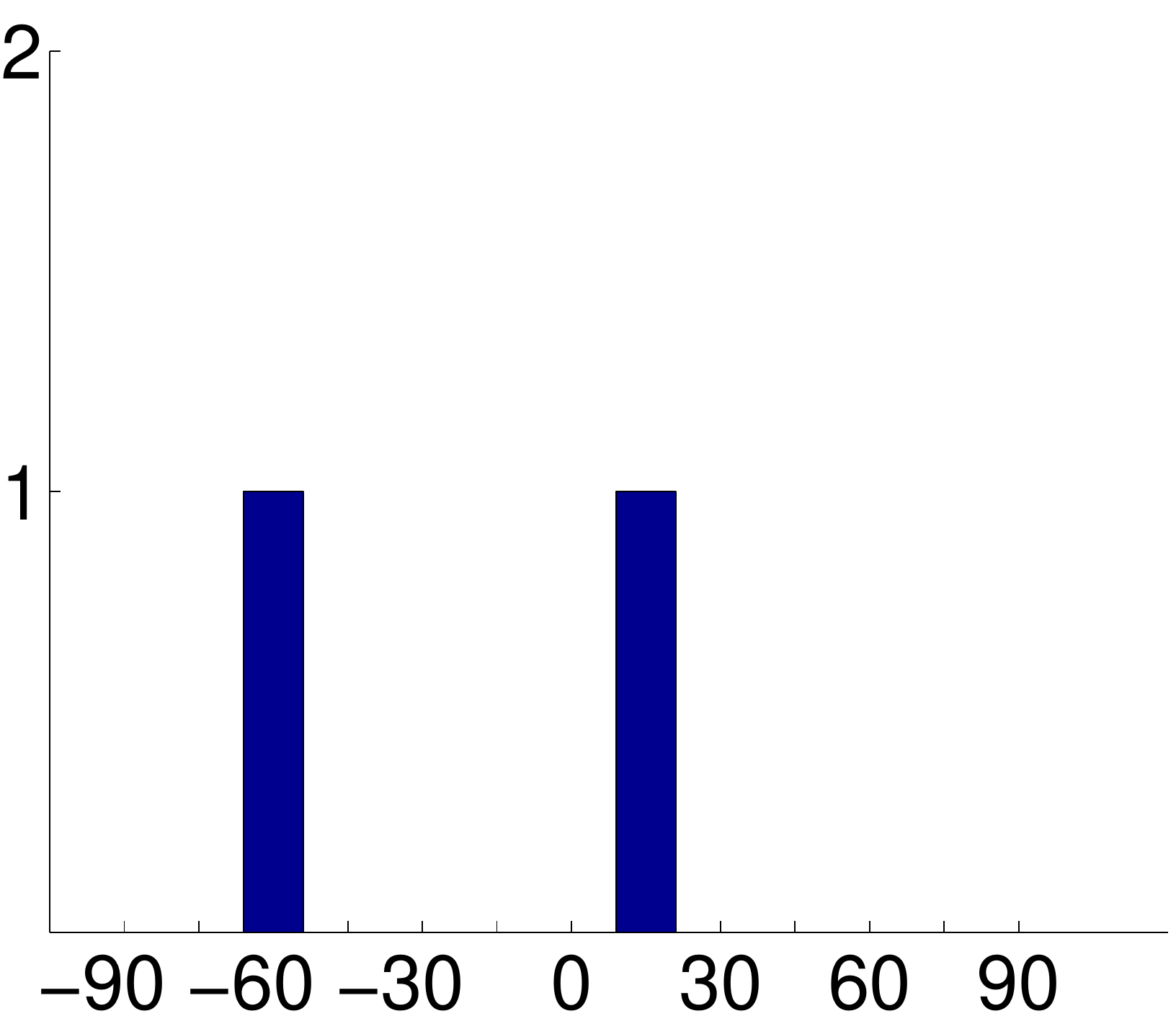}
			\includegraphics[width=0.2\linewidth,height=1.3cm]{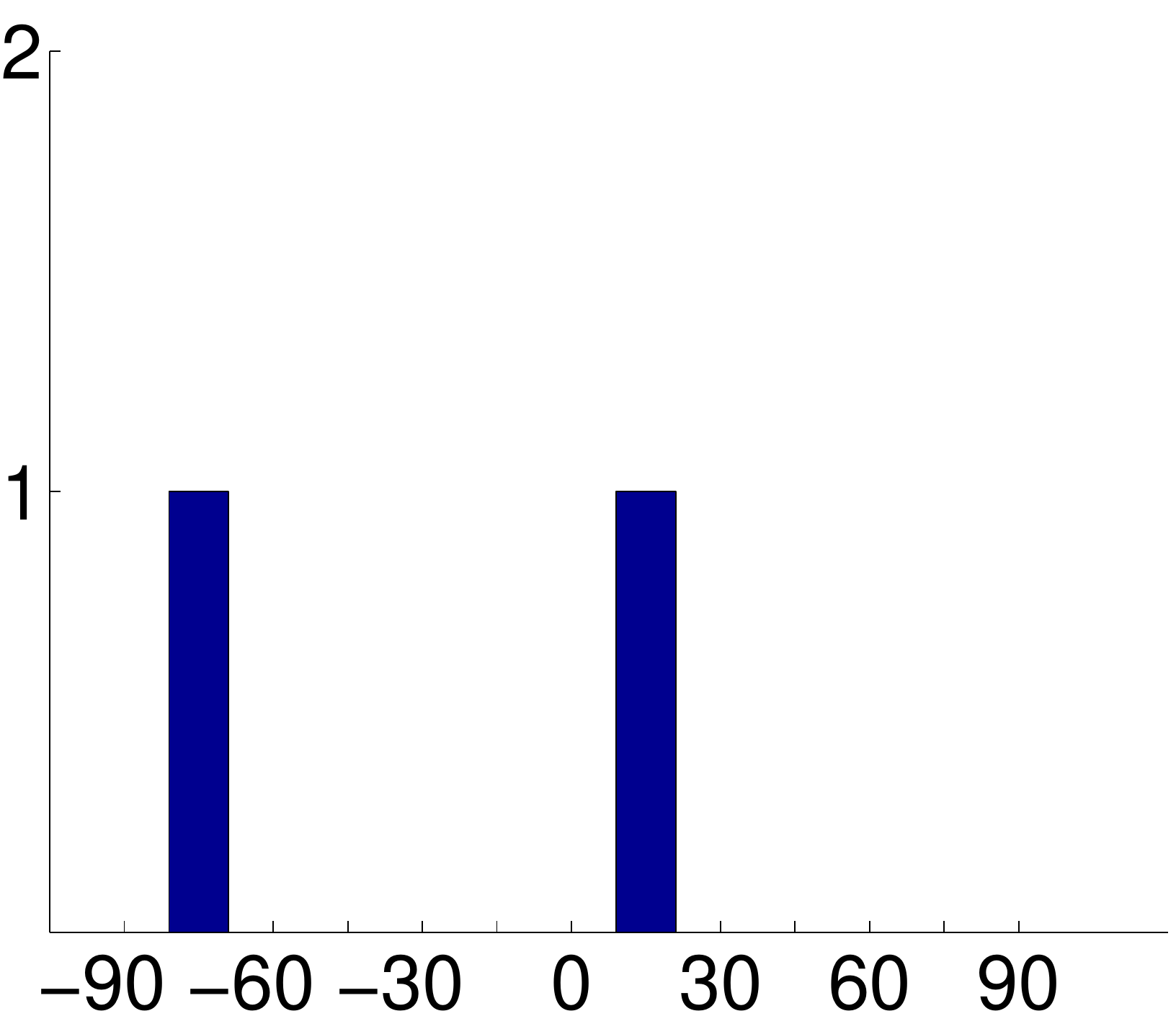}\\
		\end{tabular}
	\end{center}
		\vspace{-3mm}
		\caption{Extraction of the histogram for the orientation of faces. First row shows face detection and orientation estimation results and second row shows the histograms of face orientations (HFO) created based on these orientations. HFO has 13 dimensions which are mapped from the orientation range [-90:90] with a 15\degree intervals.}
		\label{fig:orientationHist}
	\end{figure}

\noindent\textbf{Histogram of Face Orientations (HFO):} In order to account of the distribution of the face orientations, we propose to use Histogram of Face Orientations (HFO), which simply is based on the count of face orientations for each angle in an image. In another words, it is the distribution of face orientation frequencies in an image. This descriptor has 13 feature dimensions (13 histogram bins), which corresponds to 15\degree resolution in [-90\degree - 90\degree] interval. Figure \ref{fig:orientationHist} shows some example HFO descriptors.

\vspace{2mm}

\noindent\textbf{Histogram of Face Directions (HFD):} We observe that using orientations with a lower resolution can also be useful to discriminate interactions. Based on this observation, we form a coarser histogram representation of the face orientations, and call it Histogram of face directions (HFD). This HFD feature comprises the distribution of direction frequencies in the images. Directions are defined as left, front and right which correspond to the angle intervals [-90\degree, -45\degree],  [-30\degree, 30\degree] and [45\degree, 90\degree] respectively. This descriptor is basically a coarser form of HFO and it has 3 dimensions.

\noindent\textbf{Distances of Faces (DF):} The relative locations of the faces can also be representative for an interaction. In order to capture this information, we propose to use histogram of relative distances between faces in terms of pixels. Let $L_{i}=(x_i,y_i)$ be the location of each face center and $D_{i}$ is the distance of each face to the global center of faces. If there are two faces in the image, the distance is simply the Euclidean distance of the faces. If there are more than two faces in the image, first the center point $C$ of the faces is calculated
\begin{equation}\label{eq:center}
C(x,y) =(\frac{1}{N} \sum_{i=1}^{N} L_x , \frac{1}{N} \sum_{i=1}^{N} L_y )
\end{equation}
\noindent where N is the number of faces detected in the image. To normalize the distances, we divide each of the distances to the maximum face size $S = max(s_i)$, where $s_i = max(s_x, s_y)$ is the max edge size of $i$th face. Then, $D_i$ is calculated as
\begin{equation}\label{eq:distanceNormalized}
D_i = \min(K, \max(1, \lceil \sqrt{|L_i-C|^2}/S \rceil))
\end{equation}
\noindent where, \(K\) is the maximum distance value for a face.

The final step for producing a global descriptor based on relative distances of the faces is to form the histogram of distances which simply comprises the distribution of distance frequencies in the images. In our descriptor definition, the maximum distance value $K$ defines the number of bins of the histogram since distance values are bounded to the interval $[1,K]$ and discretized into equal-sized bins. In our experiments, we choose $K=5$.

\begin{figure}
	\begin{center}
		\begin{tabular}{ccc}
			\includegraphics[width=0.2\linewidth,height=1.3cm]{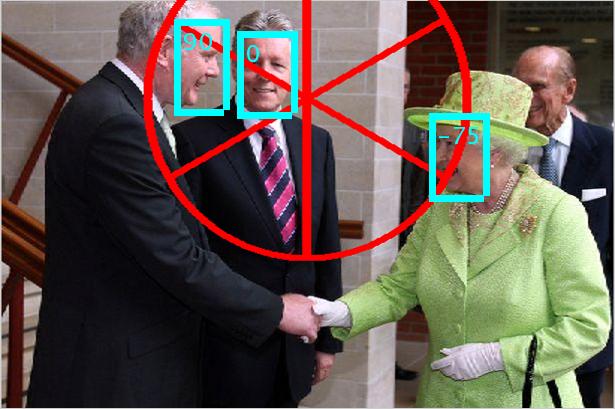}
			\includegraphics[width=0.2\linewidth,height=1.3cm]{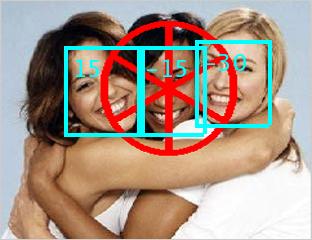}
			\includegraphics[width=0.2\linewidth,height=1.3cm]{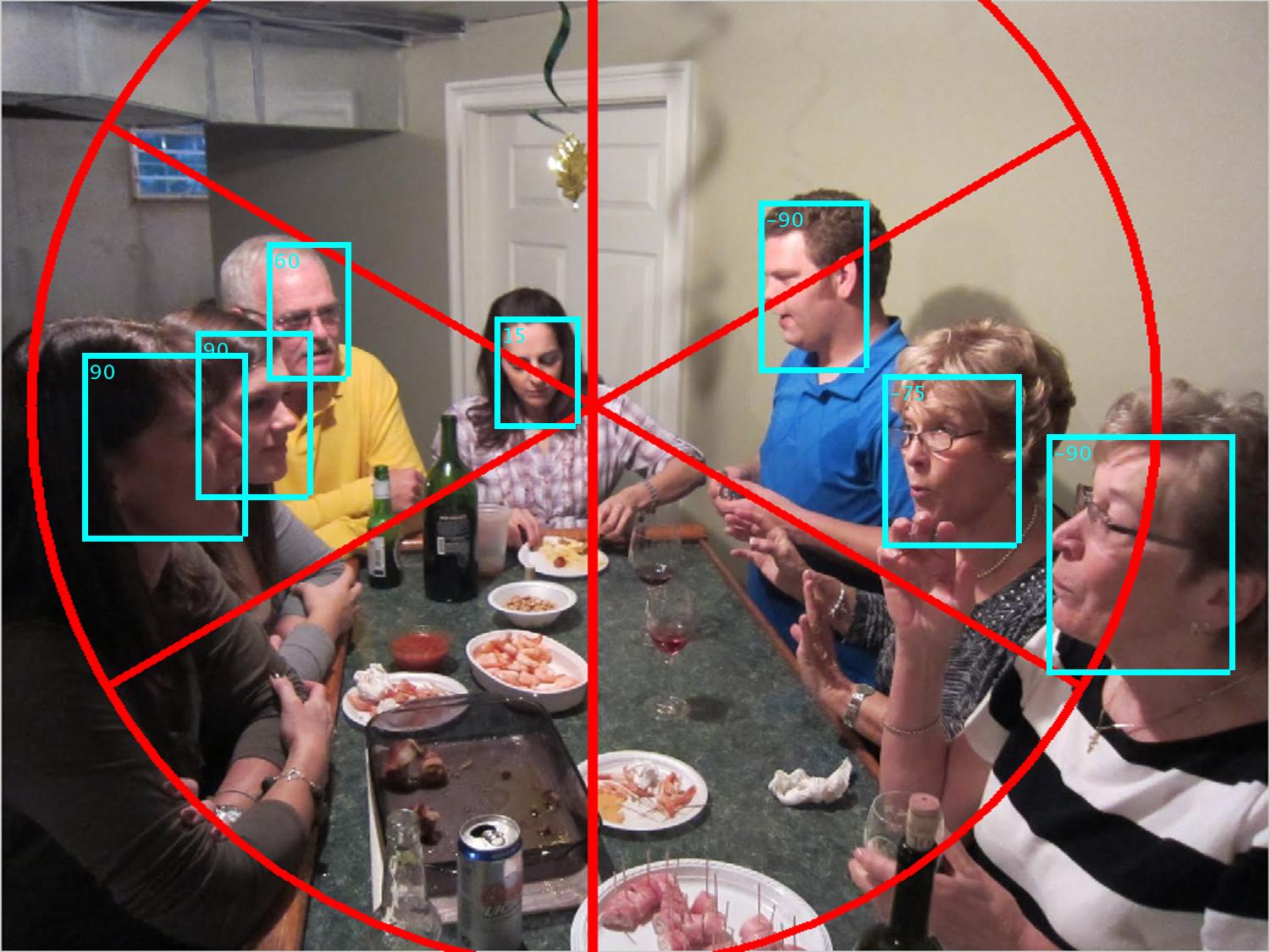}
			\includegraphics[width=0.2\linewidth,height=1.3cm]{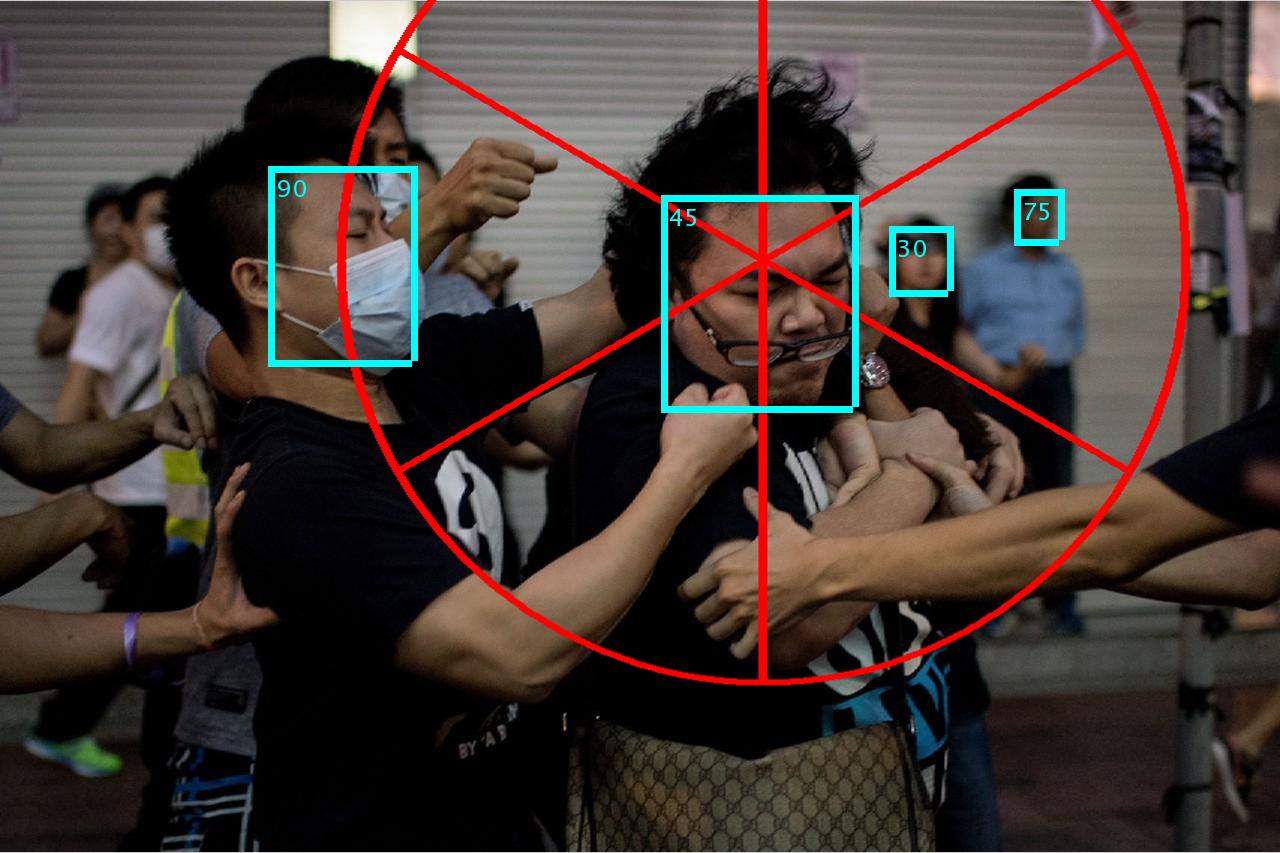}\\
			\includegraphics[width=0.2\linewidth,height=1.2cm]{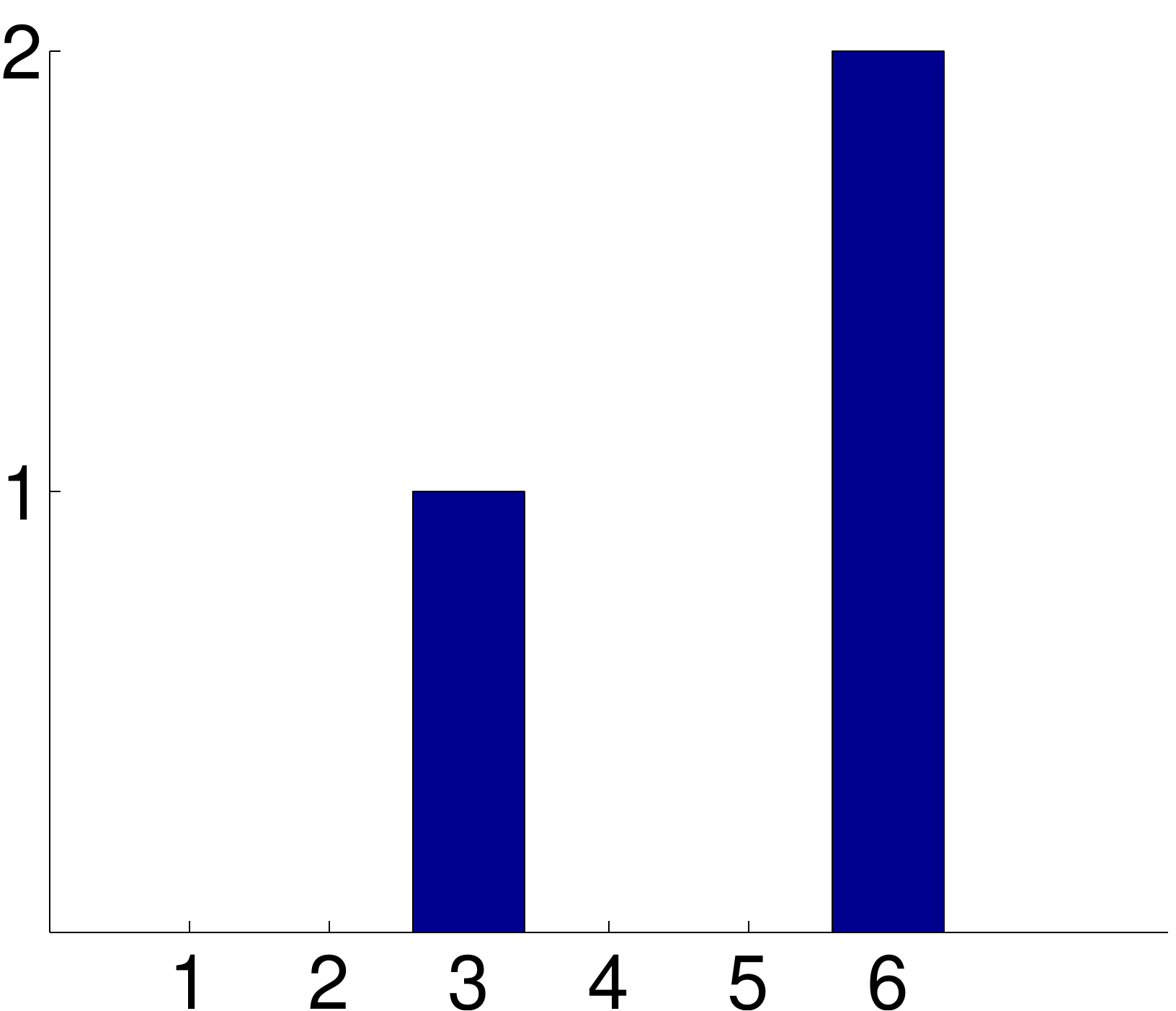}\hfill
			\includegraphics[width=0.2\linewidth,height=1.2cm]{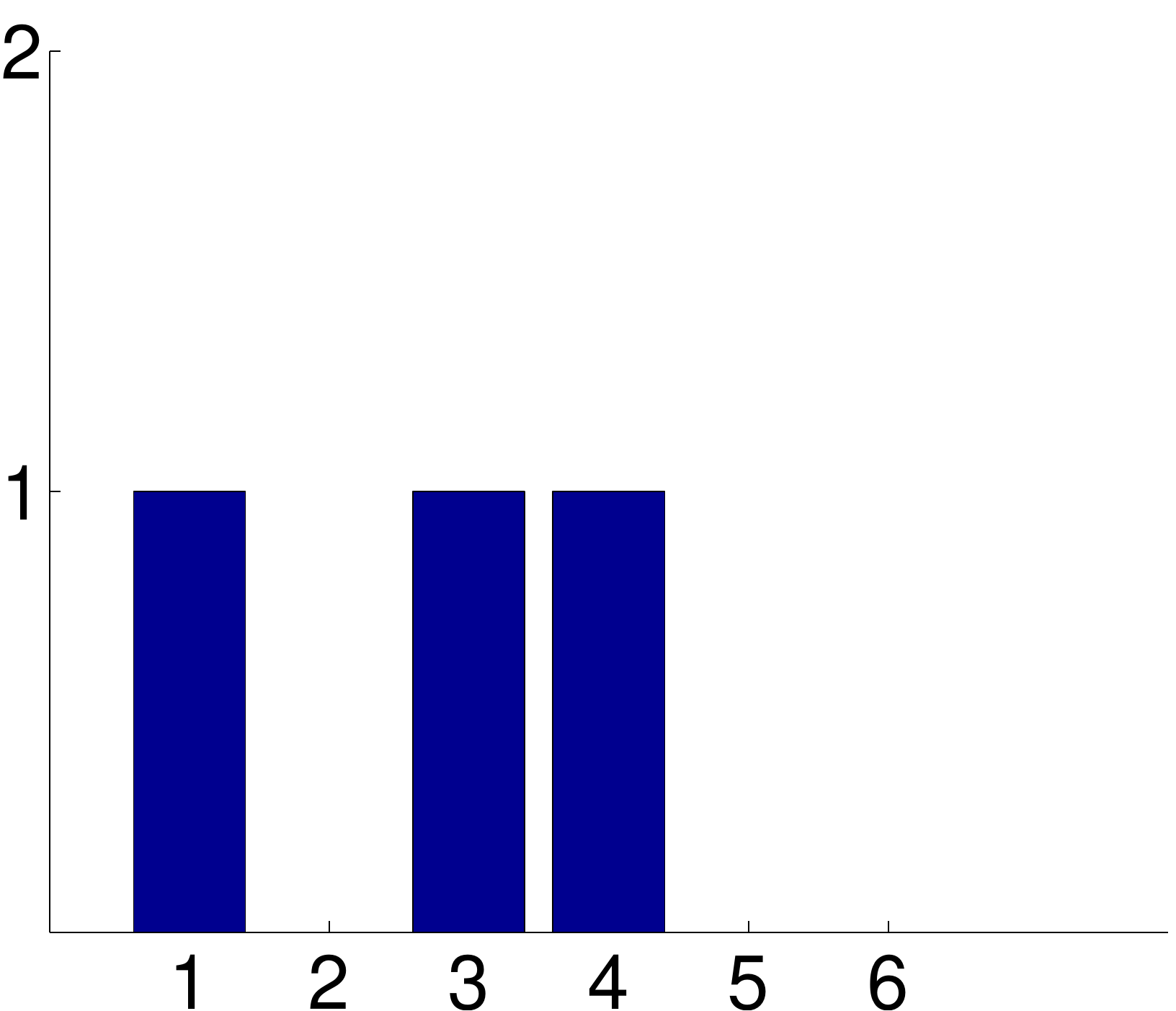}\hfill
			\includegraphics[width=0.2\linewidth,height=1.2cm]{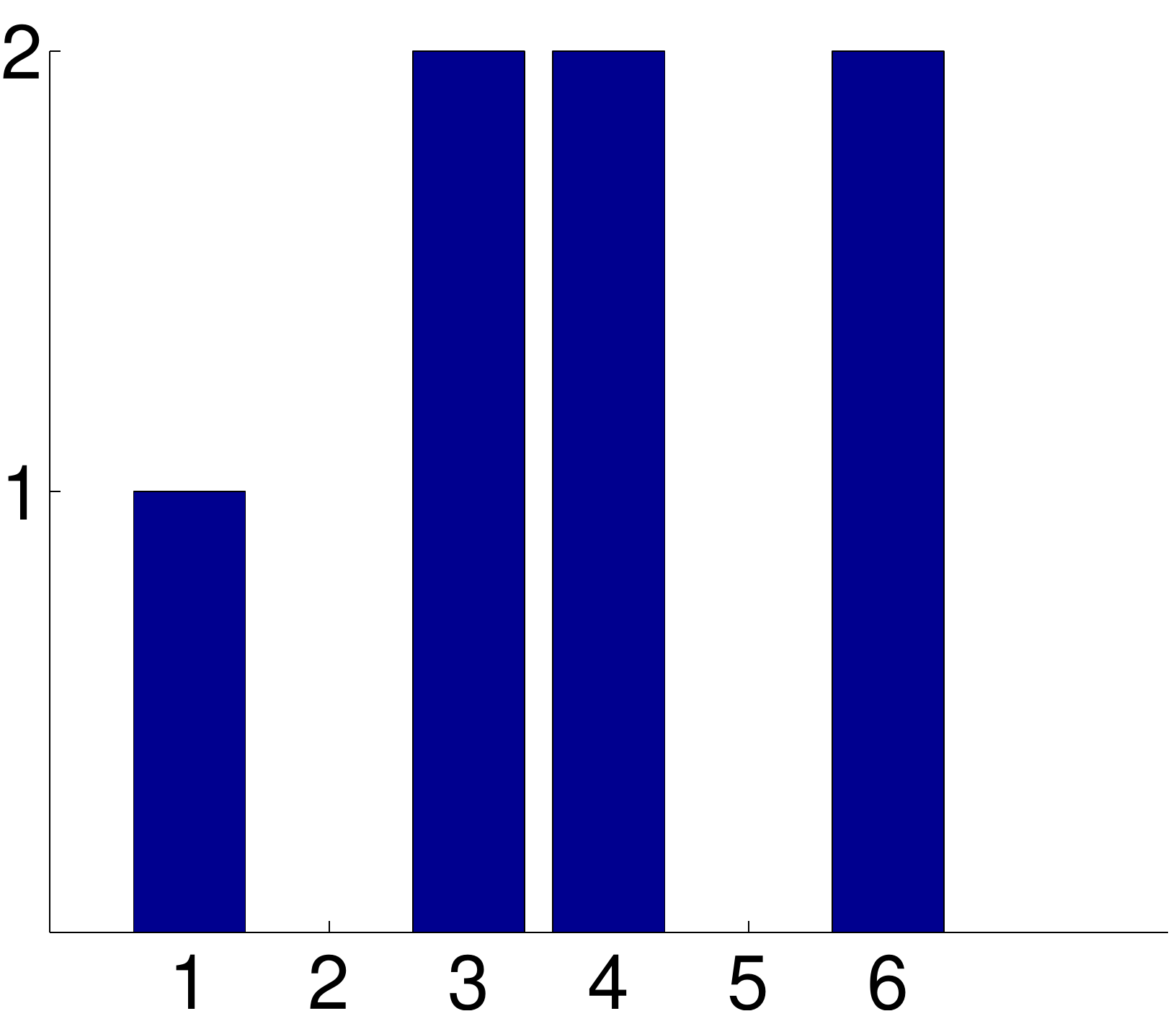}\hfill
			\includegraphics[width=0.2\linewidth,height=1.2cm]{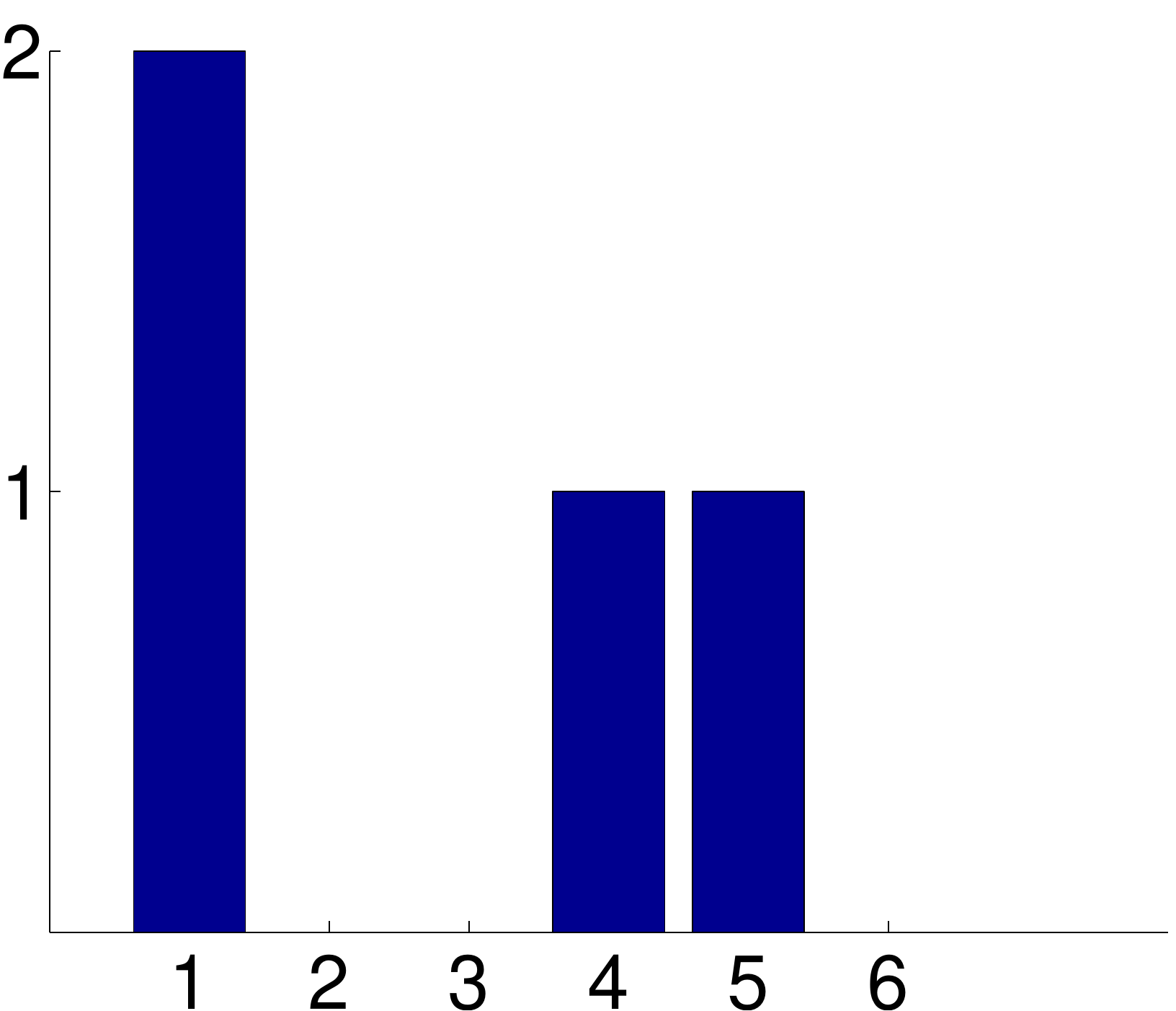}\hfill
		\end{tabular}
	\end{center}
		\vspace{-4mm}
		\caption{Example images for the calculation of Circular Histogram of Face Locations(CHFL).  First, the center of the faces and the face with the maximum distance to center is found. Then, a circle is fit on the center a with radius equal to the maximum distance that a face have to the center. Then, the circle is split into 60\degree pie regions and the distribution of faces falling into the (360\degree/60\degree) pies are calculated via a histogram. Each histogram bin holds the number of faces falling into the corresponding region.}
		\label{fig:spatialLocs}
				
	\end{figure}

\vspace{2mm}
\noindent\textbf{Circular Histogram of Face Locations (CHFL):} In order to capture the relative layout of people within an image, we propose to use a histogram of their locations. For this purpose, a circle is fit to the center of the extracted faces within an image. The center $C$ of the faces is calculated as in Eq. \ref{eq:center}. Radius of the circle is calculated as the maximum face distance to the center: $r = max(D_i)$. Thereafter, the circle is divided into equal parts like a pie chart by an angle $\alpha$. Then the distribution of faces over the pies are calculated as an histogram. To decide which face lies on which pie, the angle $\phi$ of the line that crosses both $C$ and $L_i$ is calculated as:
\begin{equation}\label{eq:angle}
\phi = \arctan{\left(\frac{y_i - C_y}{x_i - C_x}\right)}
\end{equation}
\noindent where $x_i$ and $y_i$ are the x-y coordinates of $i$th face in the image. having identified the pie index of face in the circle, the last step is to create a histogram which comprises the distribution of face frequencies over the pies of the circle. The histogram consists of $2\pi / \alpha$ bins, and the location of the face projected on the histogram as $ \alpha / \phi$. Figure \ref{fig:spatialLocs} shows example cases on how this CHFL global descriptor is extracted.

\vspace{2mm}
\noindent\textbf{Grid Histogram of Face Locations (GHFL):} Similar to CHFL, we also form the grid histogram of face locations, in order to capture the spatial layout of the multiple people within an image. For this purpose, the image is split into the $M \times N$ size grids where the center of the middle grid is the center of the faces. A $M \times N$-bin histogram is created and the number of faces falling within the respective grid is computed. Our preliminary results show that $1\times 3$ grid size gives the best results, since for most of the interaction images, faces lie on the horizontal plane. 

In addition to the visual features described above, we also include the number of different facial orientation directions to our list of global facial descriptors. Our final feature length for the combined descriptor therefore becomes 31.
   
\vspace{2mm}
\noindent\textbf{Scene features:} In order to capture the general characteristics of the scene and to investigate its influence on human-human interaction recognition, we use GIST descriptors proposed by \cite{gist2001}. GIST features provide a low-dimensional representation of the global scene layout. 

In addition, we utilize Bag-of-Words (BoW), both as a baseline and a global scene descriptor. For this purpose, we first extract dense SIFT features from the images and create a codebook using k-means (k=1000). Then, each image is represented with the histogram of codewords. We additionally employ spatial pyramid matching (SPM) of \cite{SPM}, with $2 \times 2$ grids.

\vspace{2mm}
\noindent\textbf{Deep features:} 
State-of-the-art in many computer vision classification tasks has received a significant gain in performance with the introduction of deep learning and the Convolutional Neural Network (CNN) architectures. 

In order to explore the effect of these architectures over the human interaction image recognition problem, we make use of two kinds of deep features, which are trained using the large scene dataset called Places (\cite{CNN}). This deep feature is called Places-CNN and is trained using the Caffe package over the dataset of  2,448,873 images of 205 categories. The output of the FC7 layer, which is 4096-dimensional, is taken as the deep feature and a linear SVM is trained using our training set. 

We also evaluate the similar feature called Hybrid-CNN, based on the same architecture, but trained over the larger dataset that combines both the Places and the ImageNet datasets (\cite{CNN}). This joint training set has 3.5 million images from 1183 categories. Due to the broadness of classes in the training set, this Hybrid-CNN feature is likely to capture both the scene and the object characteristics. The results of using both of these deep features are given in the Experimental Evaluation section.

\section{Experimental Evaluation}
\subsection{Dataset}
\begin{figure}
\begin{center}
\includegraphics[width=0.15\linewidth,height=1.22cm]{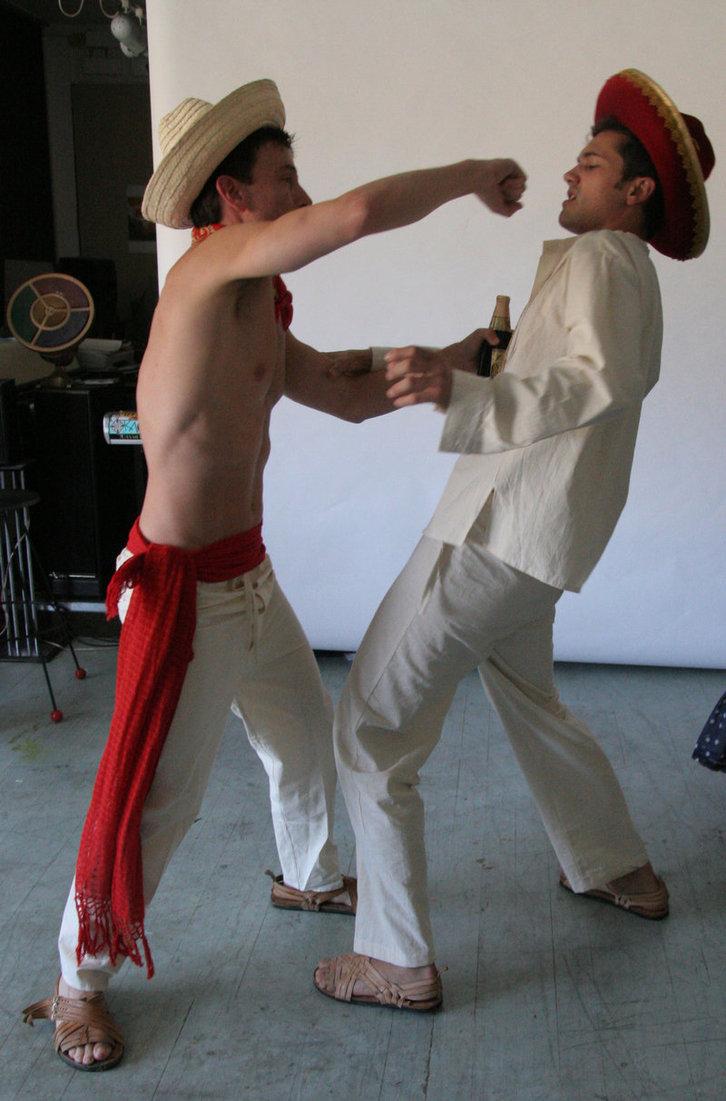}
\includegraphics[width=0.15\linewidth,height=1.22cm]{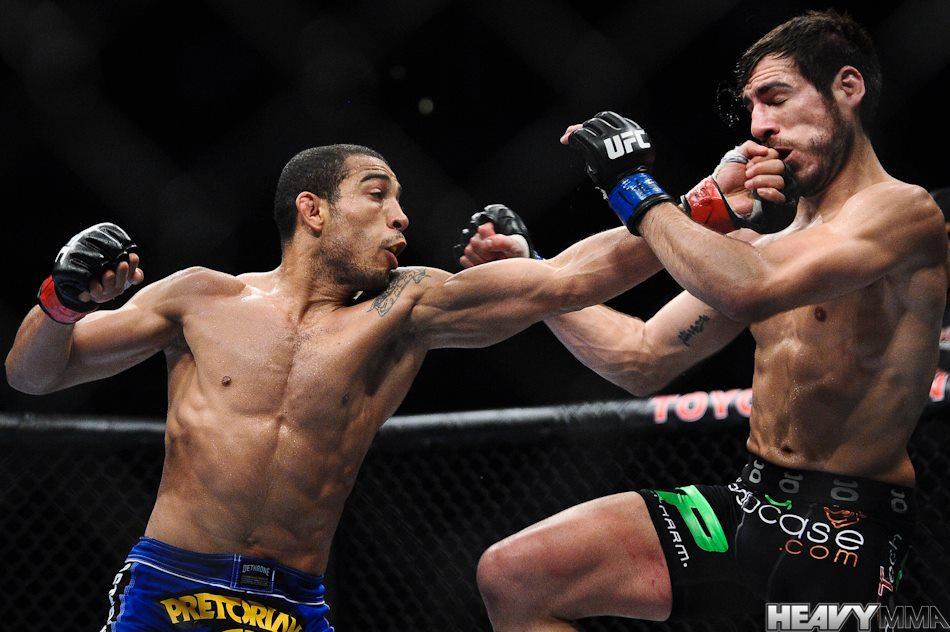}
\includegraphics[width=0.15\linewidth,height=1.22cm]{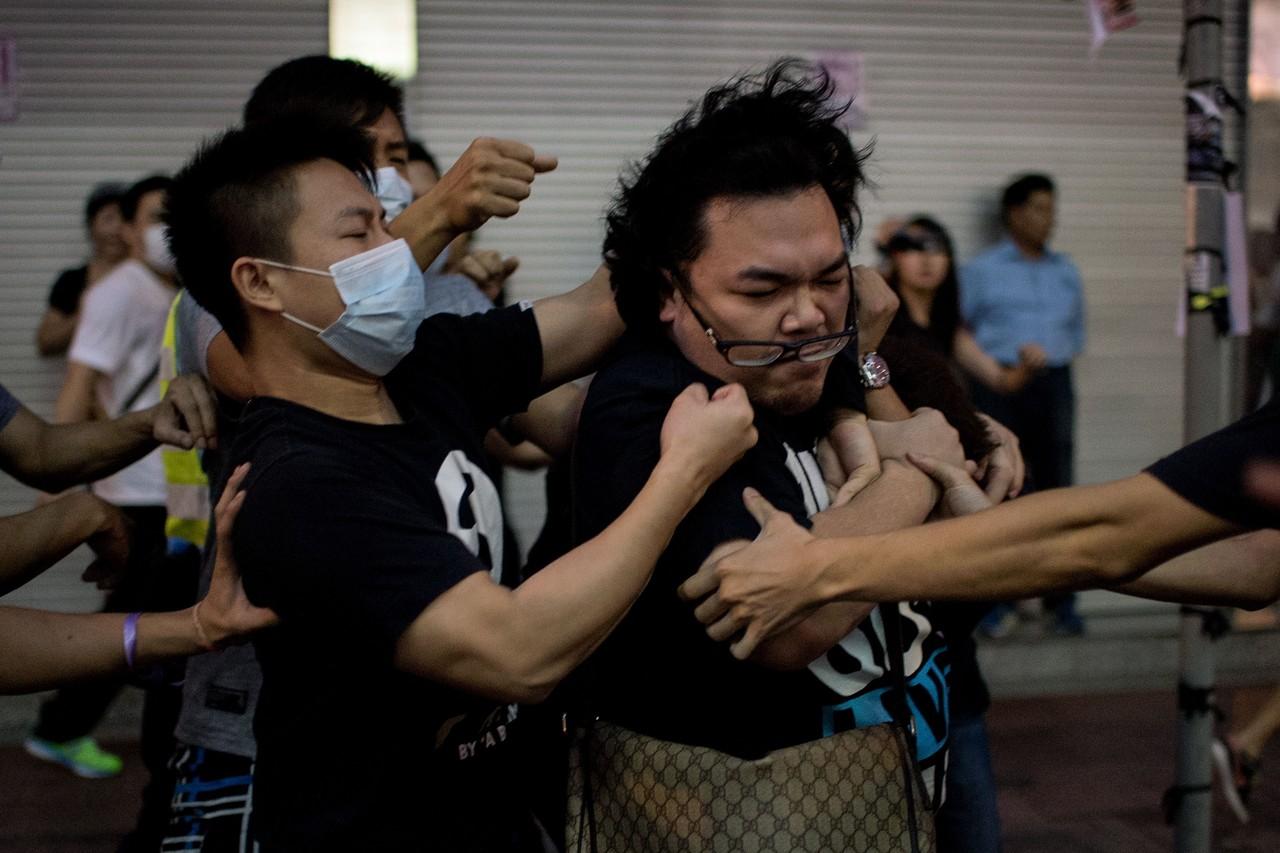}\hspace{0.1cm}
\includegraphics[width=0.15\linewidth,height=1.22cm]{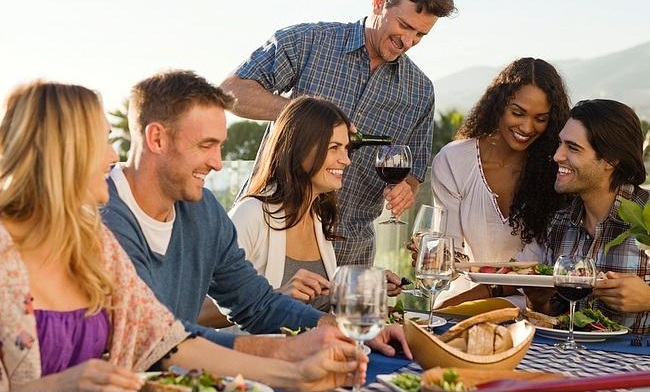}
\includegraphics[width=0.15\linewidth,height=1.22cm]{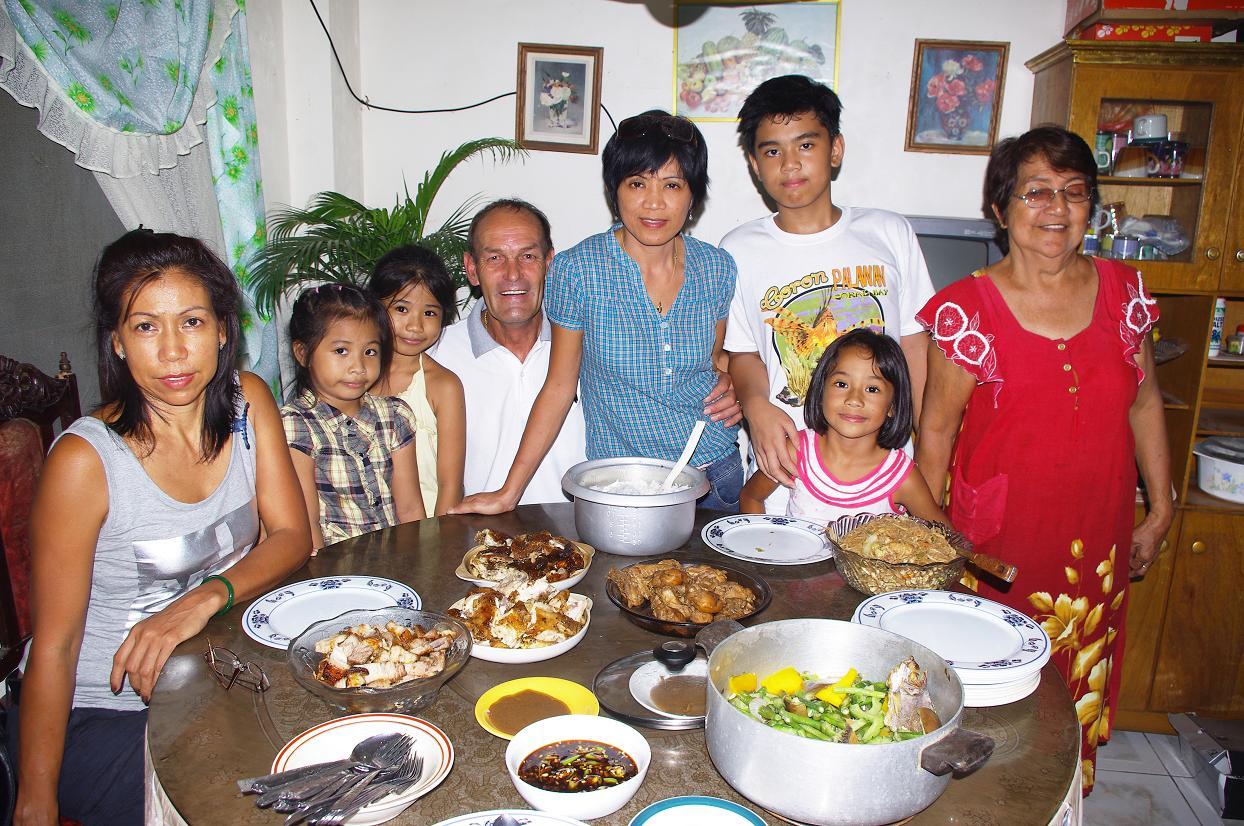}
\includegraphics[width=0.15\linewidth,height=1.22cm]{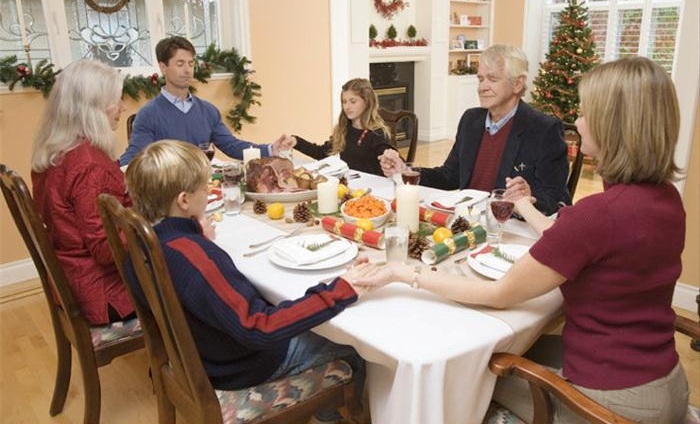} \\
\includegraphics[width=0.15\linewidth,height=1.22cm]{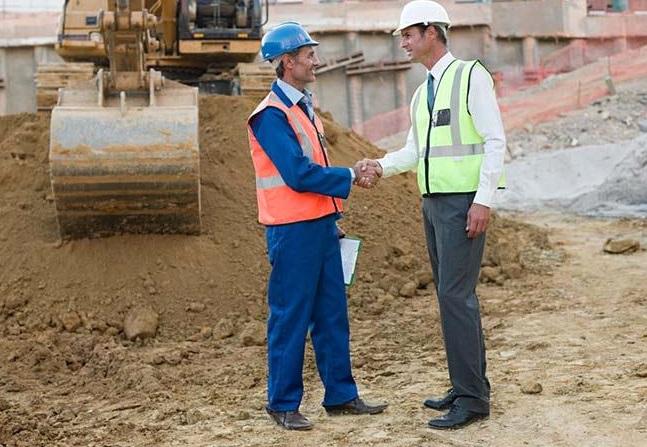}
\includegraphics[width=0.15\linewidth,height=1.22cm]{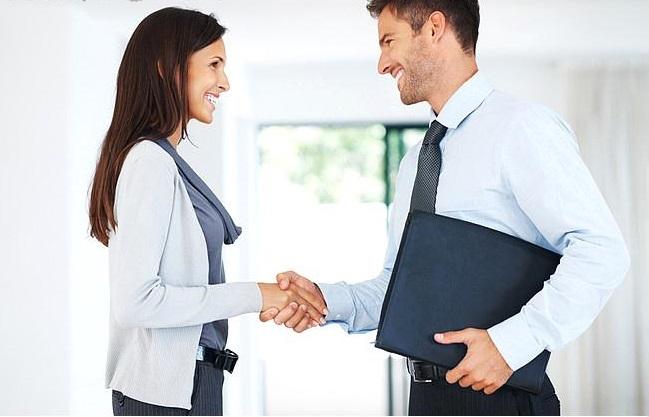}
\includegraphics[width=0.15\linewidth,height=1.22cm]{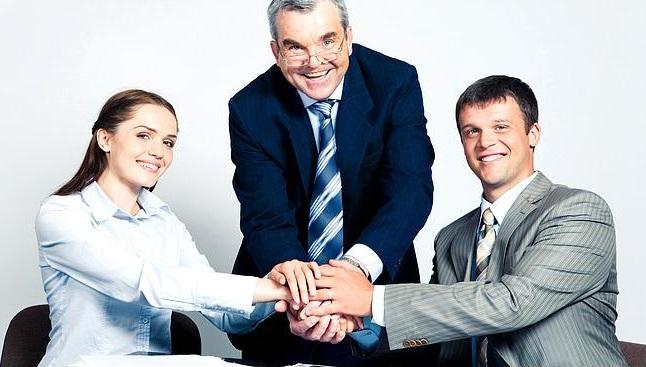}\hspace{0.1cm}
\includegraphics[width=0.15\linewidth,height=1.22cm]{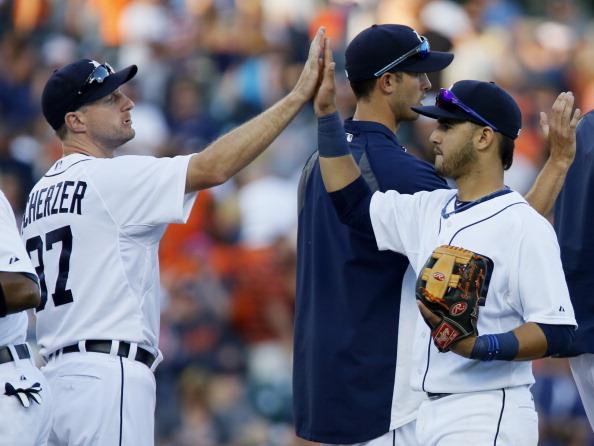}
\includegraphics[width=0.15\linewidth,height=1.22cm]{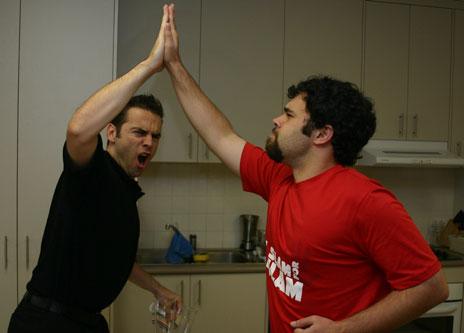}
\includegraphics[width=0.15\linewidth,height=1.22cm]{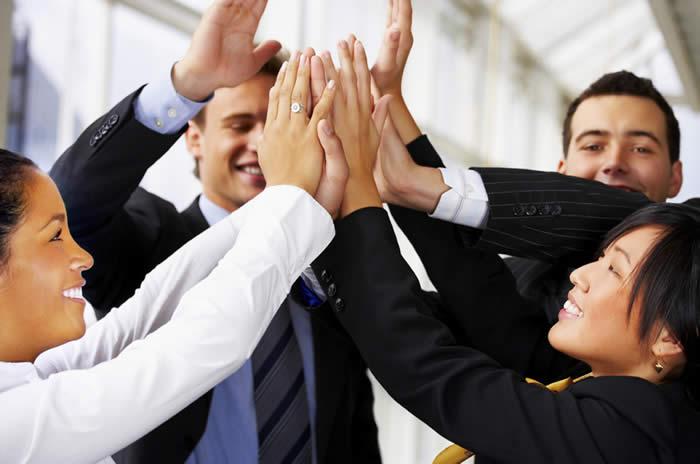}  \\
\includegraphics[width=0.15\linewidth,height=1.22cm]{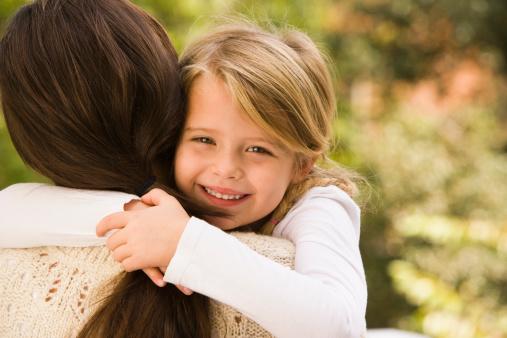}
\includegraphics[width=0.15\linewidth,height=1.22cm]{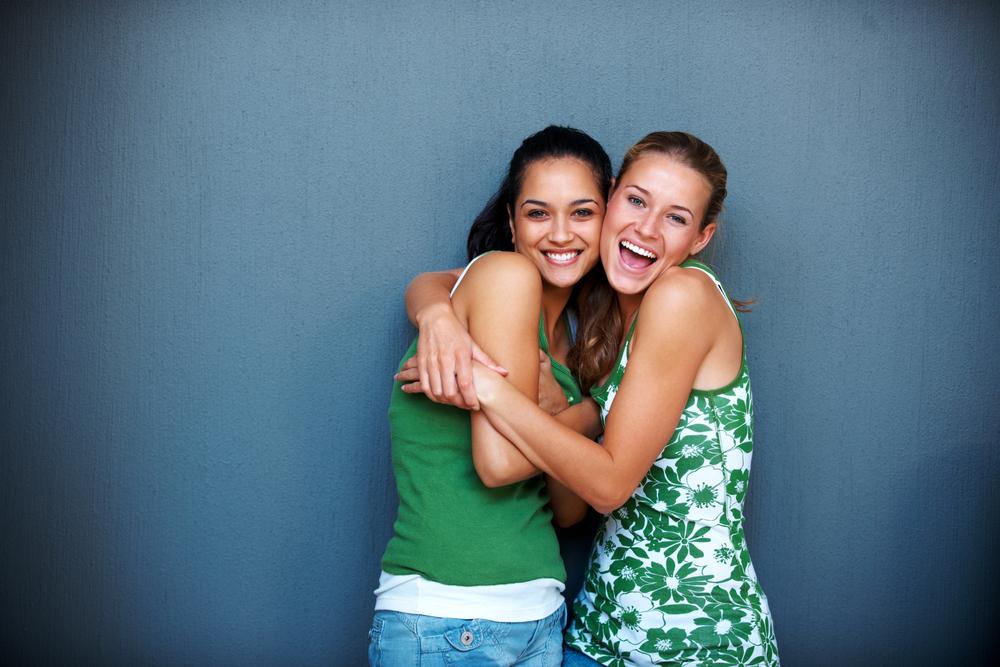}  
\includegraphics[width=0.15\linewidth,height=1.22cm]{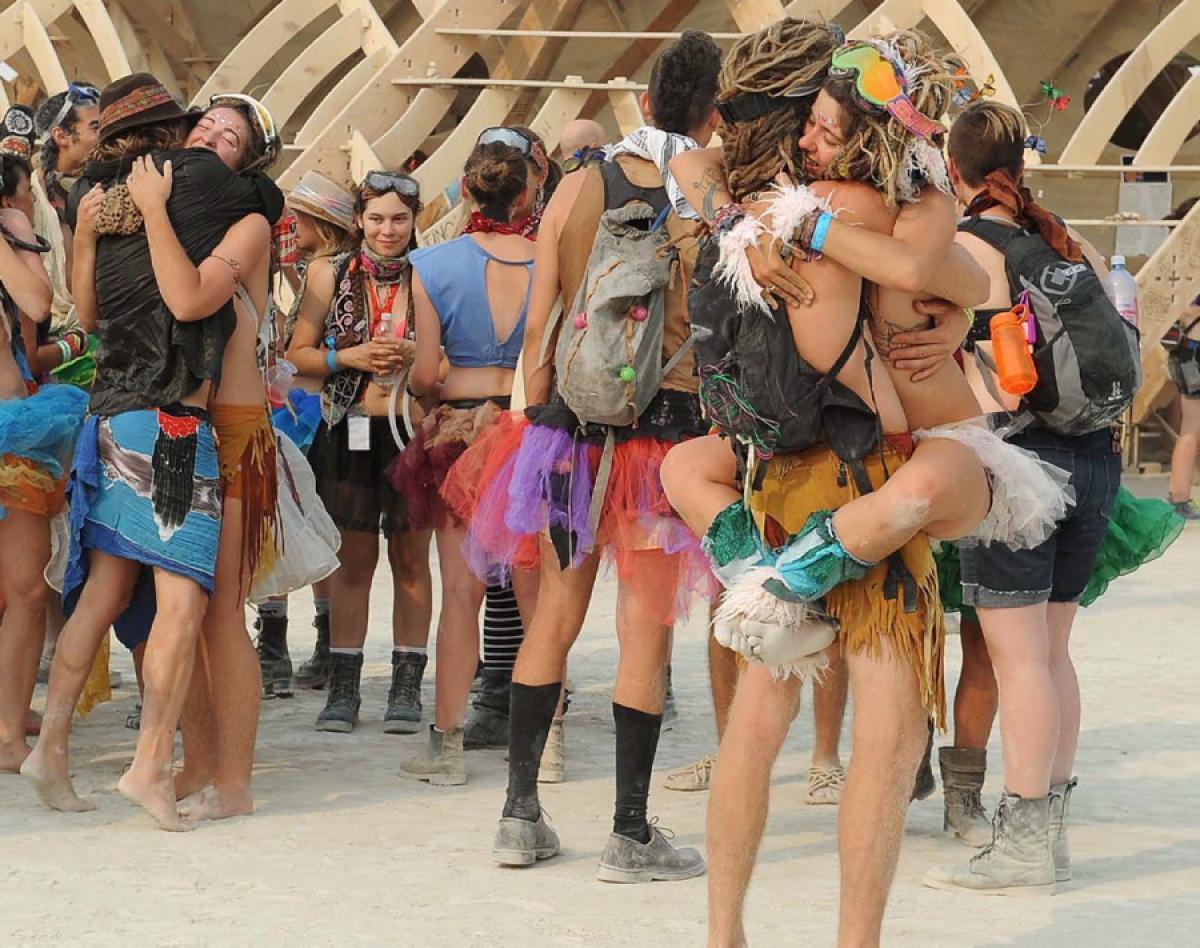} \hspace{0.1cm}
\includegraphics[width=0.15\linewidth,height=1.22cm]{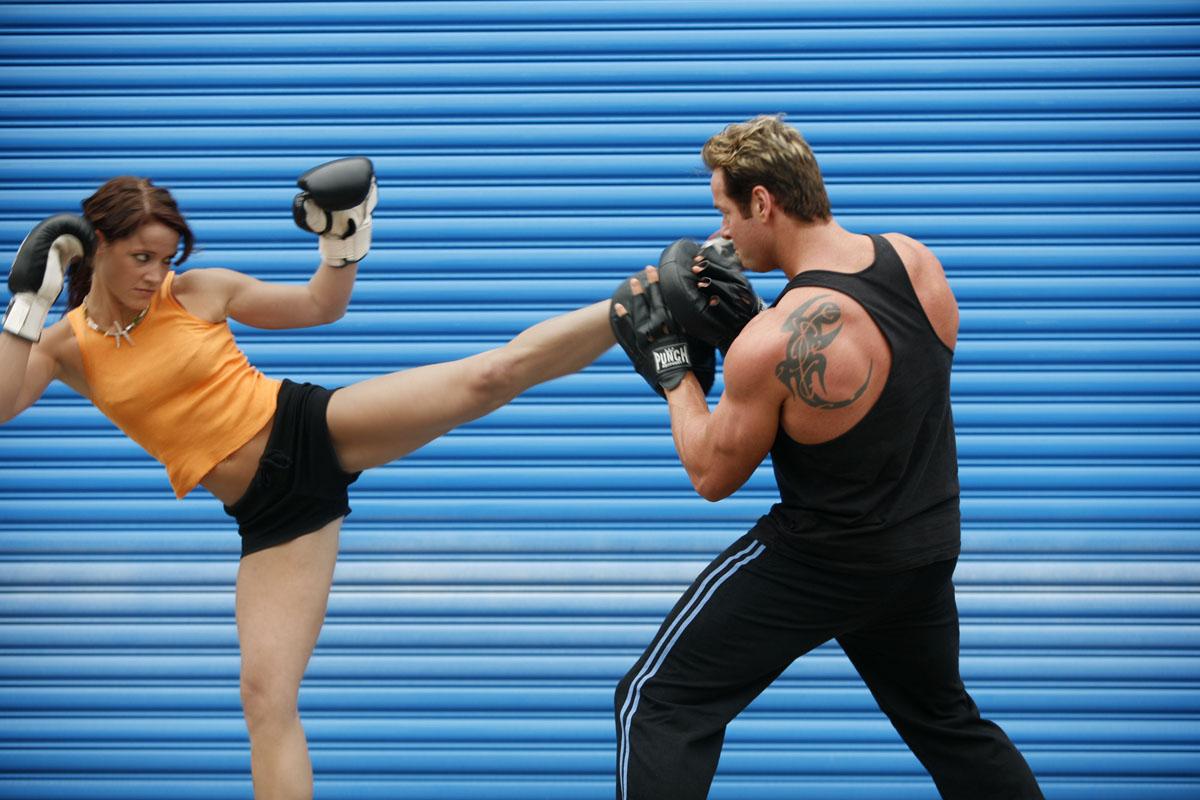}
\includegraphics[width=0.15\linewidth,height=1.22cm]{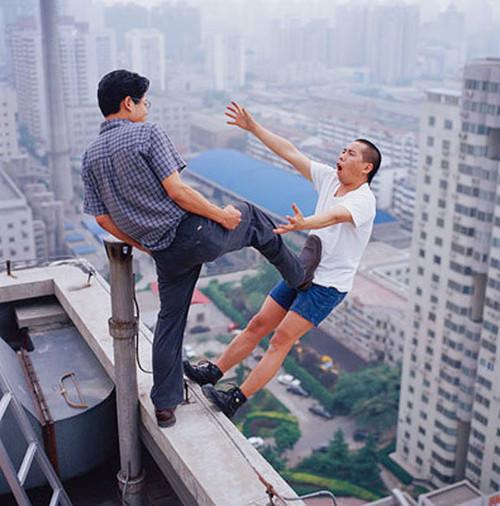} 
\includegraphics[width=0.15\linewidth,height=1.22cm]{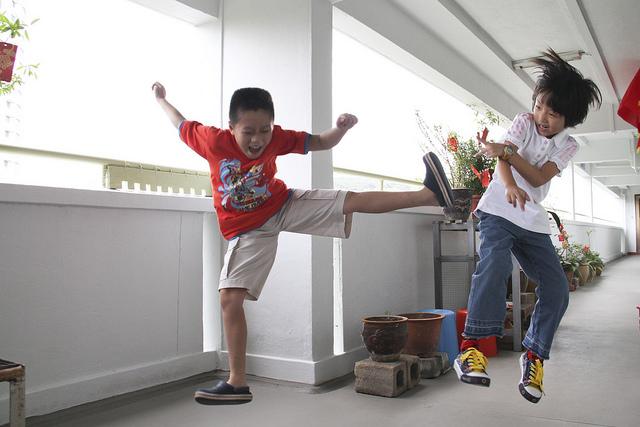} \\  
\includegraphics[width=0.15\linewidth,height=1.22cm]{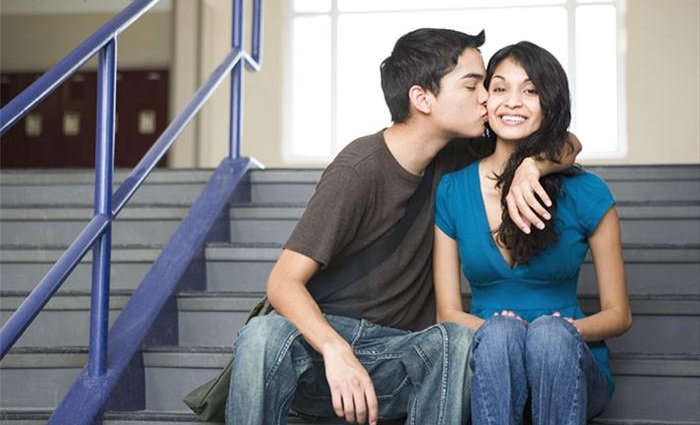}
\includegraphics[width=0.15\linewidth,height=1.22cm]{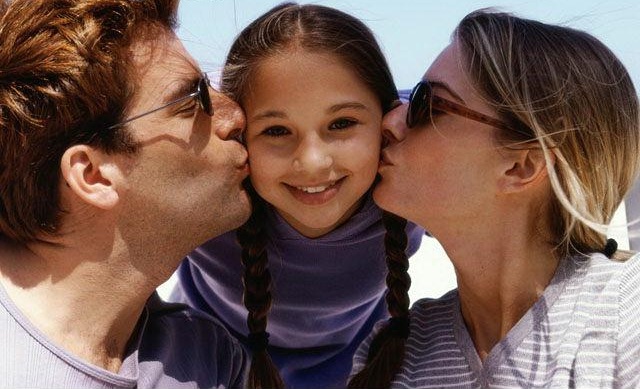}
\includegraphics[width=0.15\linewidth,height=1.22cm]{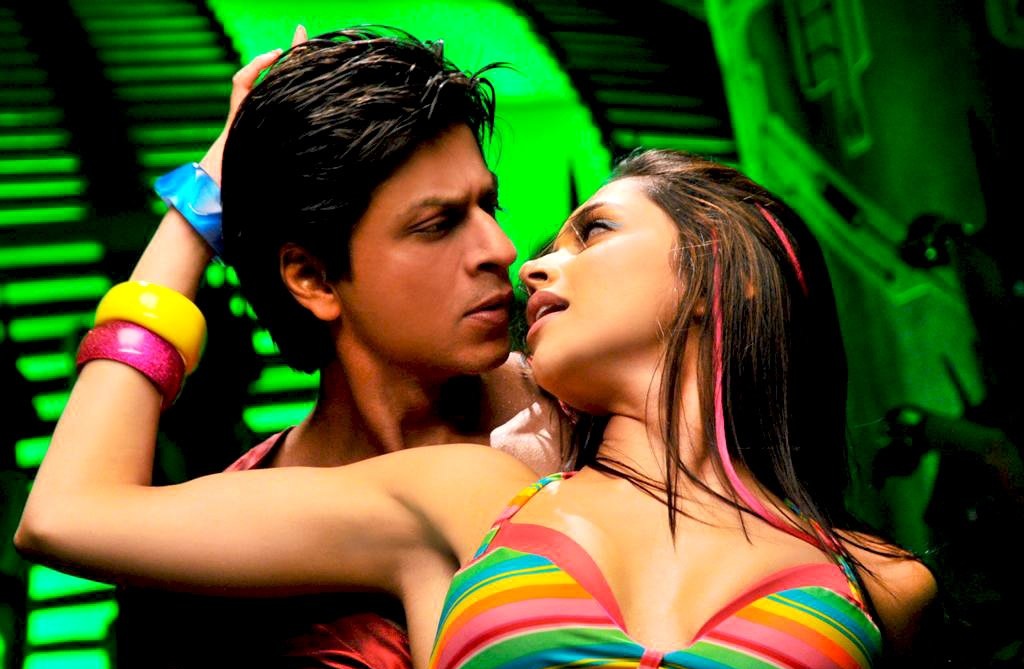} \hspace{0.1cm}
\includegraphics[width=0.15\linewidth,height=1.22cm]{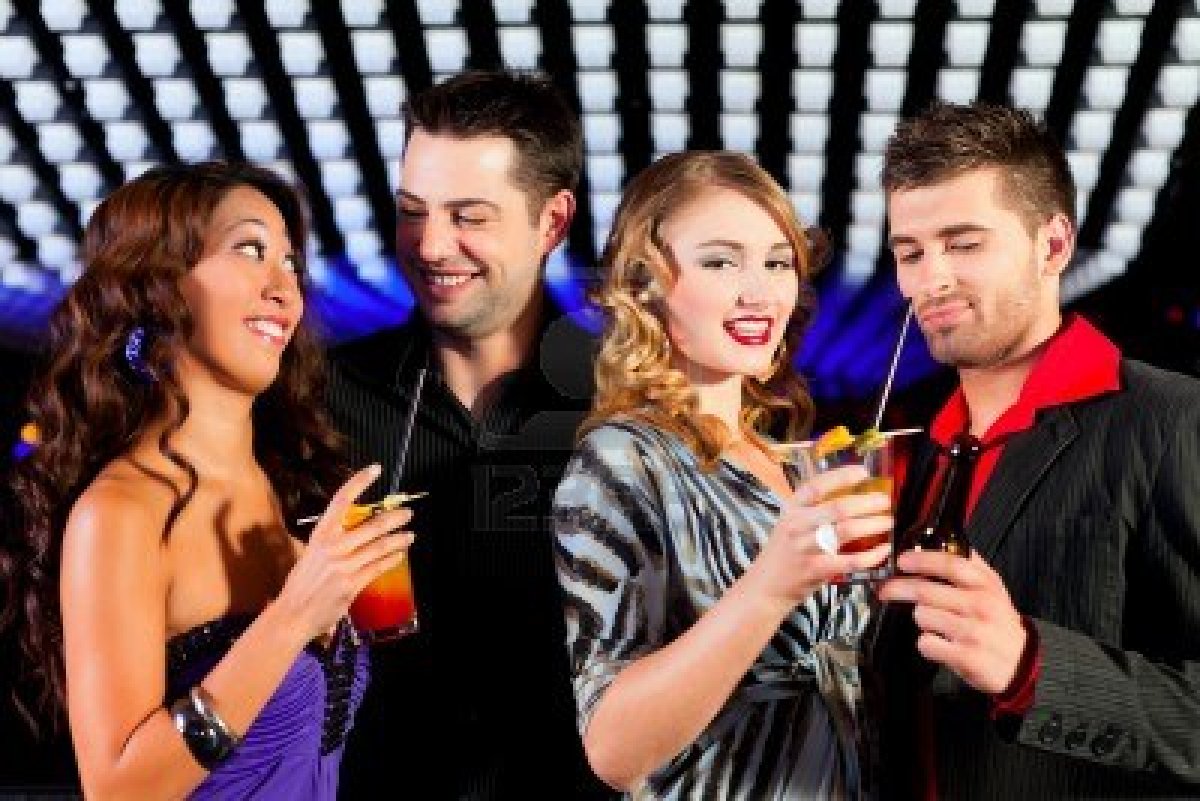}
\includegraphics[width=0.15\linewidth,height=1.22cm]{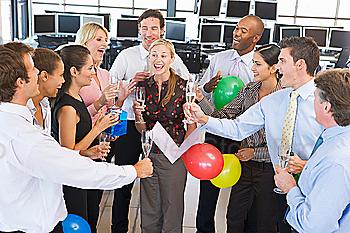} 
\includegraphics[width=0.15\linewidth,height=1.22cm]{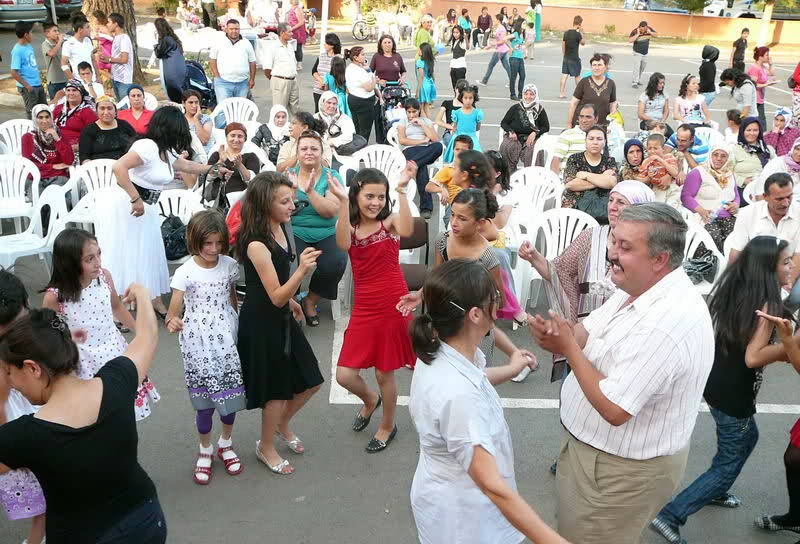}\\
\includegraphics[width=0.15\linewidth,height=1.22cm]{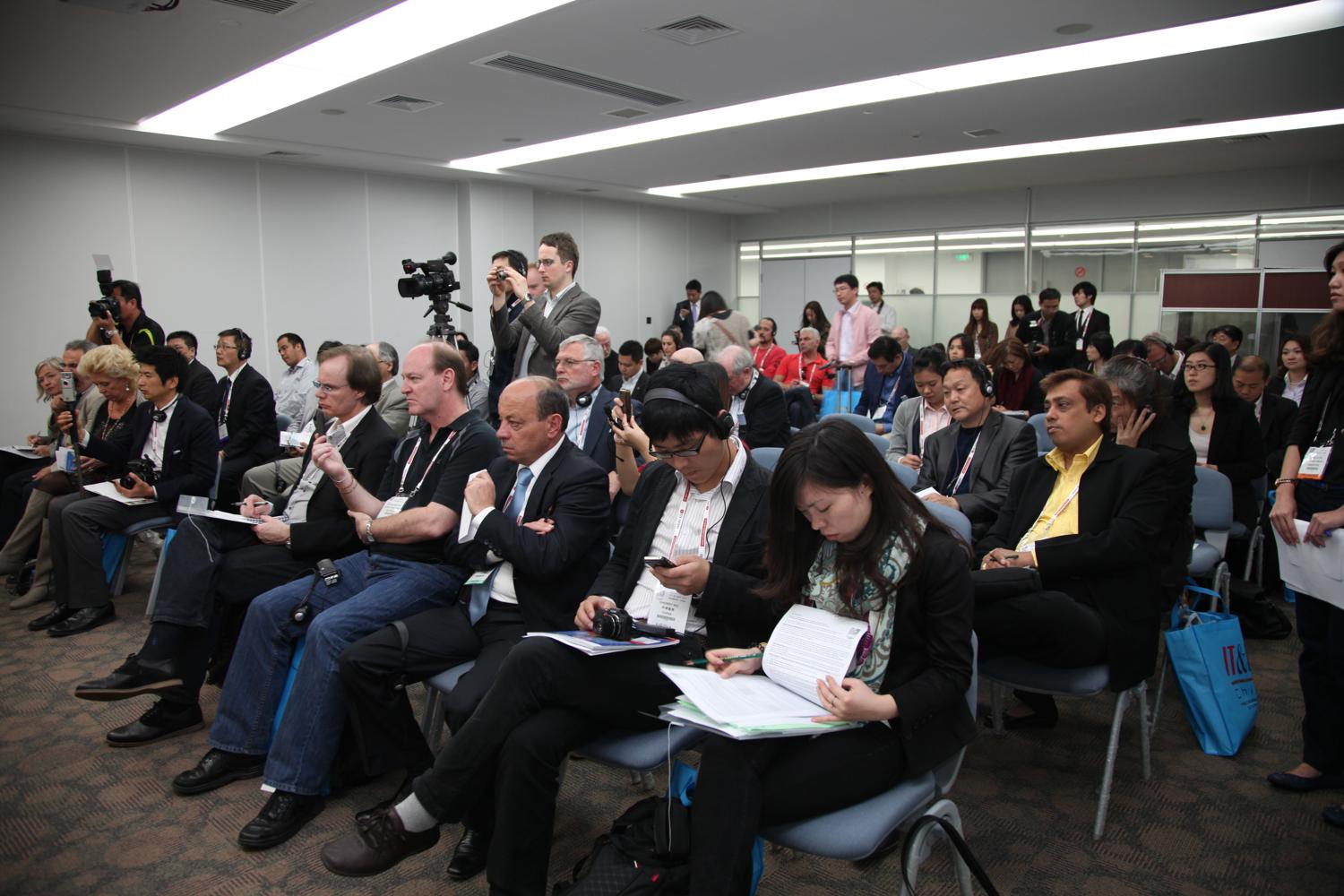}
\includegraphics[width=0.15\linewidth,height=1.22cm]{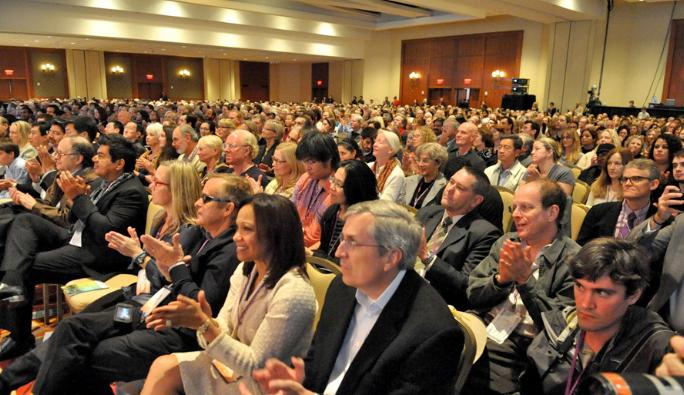}
\includegraphics[width=0.15\linewidth,height=1.22cm]{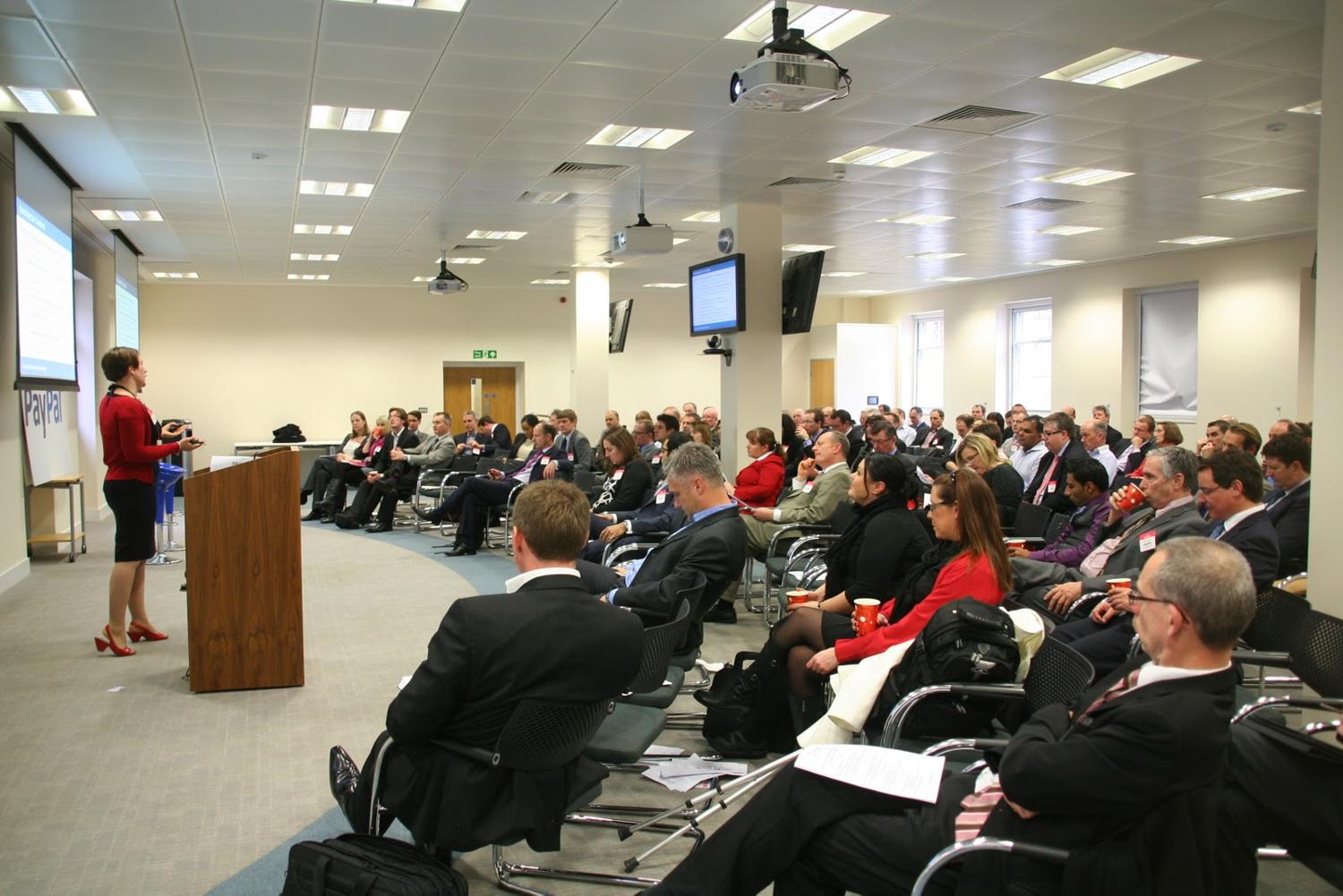} \hspace{0.1cm}
\includegraphics[width=0.15\linewidth,height=1.22cm]{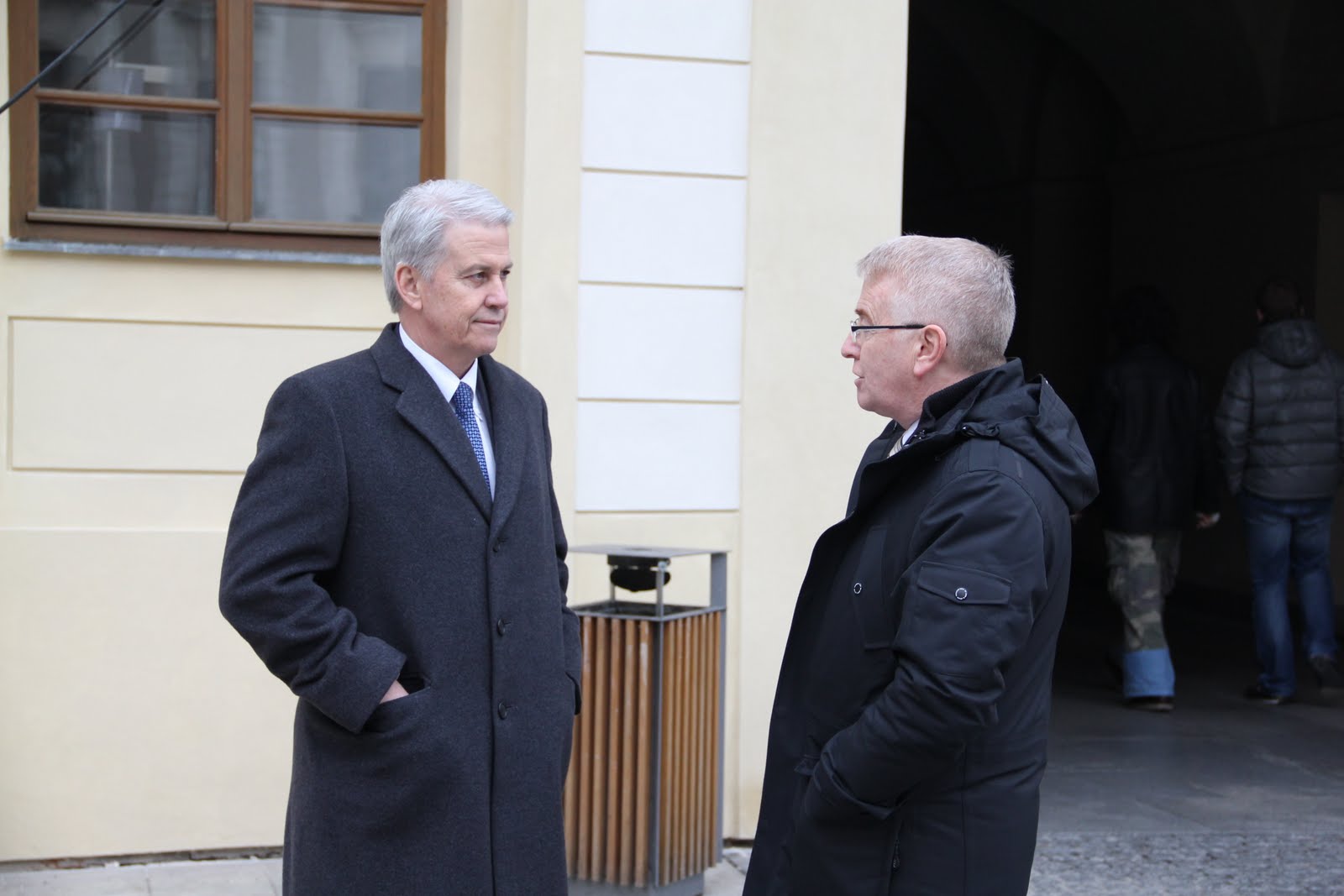}
\includegraphics[width=0.15\linewidth,height=1.22cm]{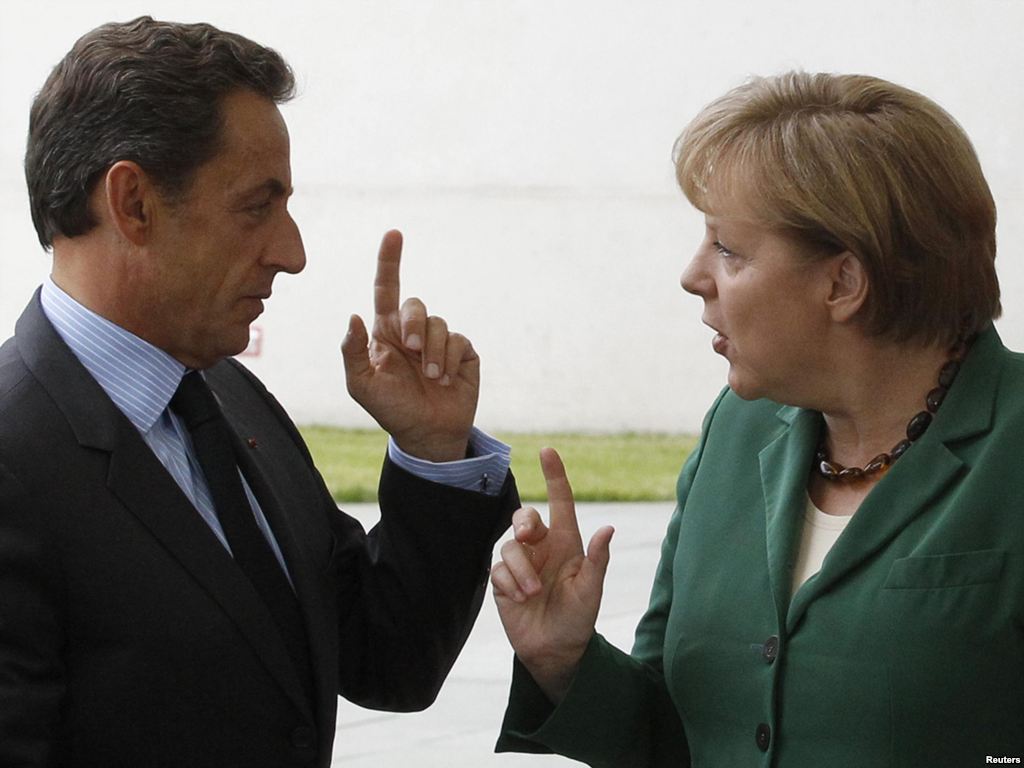}
\includegraphics[width=0.15\linewidth,height=1.22cm]{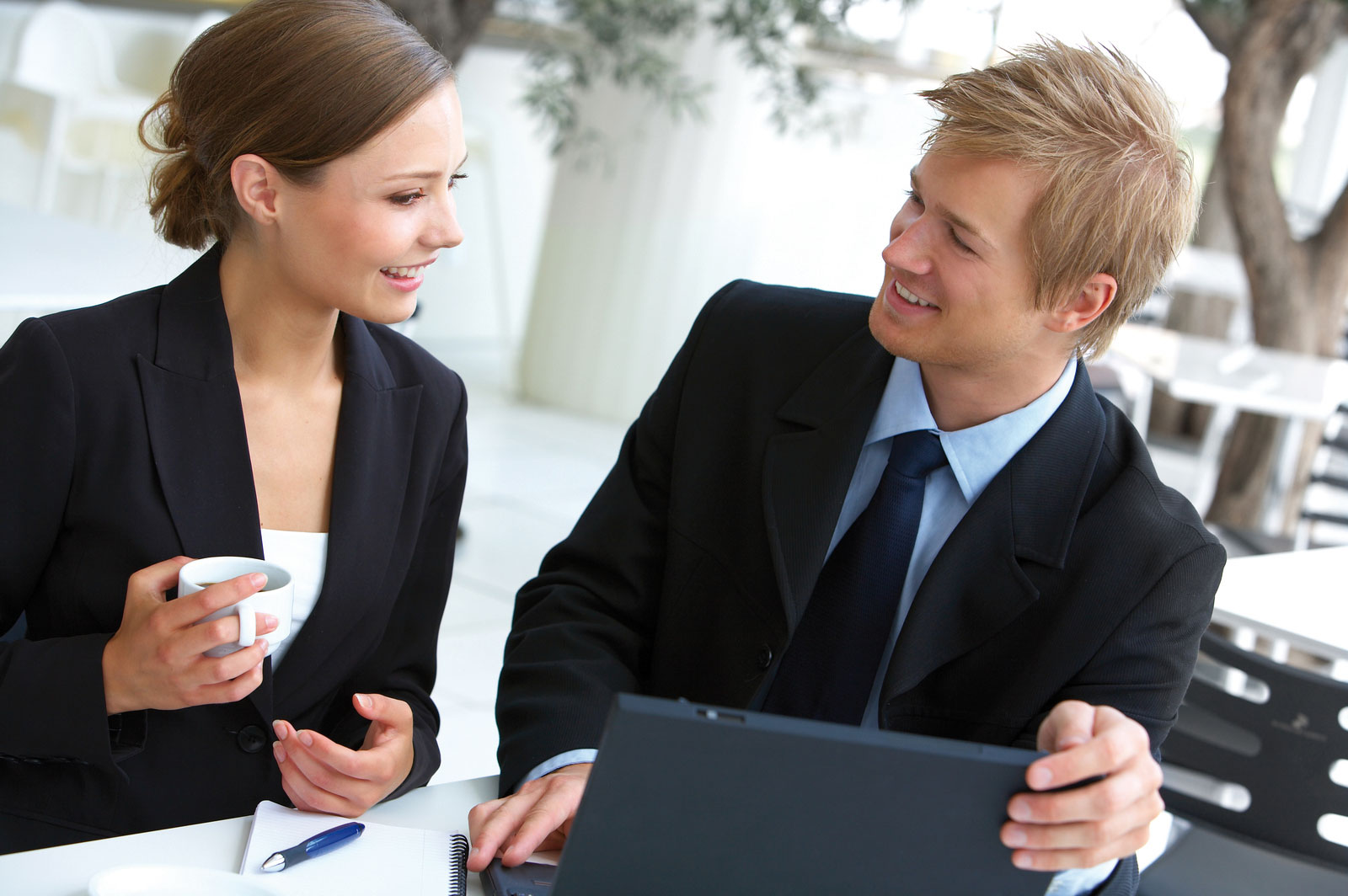} \\
\end{center}
		\vspace{-6mm}
	\caption{ Example images from the newly collected dataset of human-human still image interactions. Three example images are shown for each of the ten classes for this dataset, {\em boxing-punching, dining, handshaking, highfive, hugging, kicking, kissing, partying, speech} and {\em talking}. Note that, the poses and appearances of the people in interaction is quite diverse, making it a challenging dataset. }
\label{fig:dataset}
\end{figure}

In order to evaluate the proposed facial descriptors and their effect on human-human interaction recognition, we collected a new image dataset that includes ten human interaction classes. These classes are {\em boxing-punching, dining, handshaking, highfive, hugging, kicking, kissing, partying, speech} and {\em talking}. Each class contains at least 150 images, forming a total of 1971 images. When collecting the dataset, we gather images such that one of the target interaction classes is present and at least one person has a visible facial region in each image. The images for the {\em boxing-punching, handshaking, highfive, hugging, kicking, kissing} and {\em talking} classes usually include two to three people, whereas the number of people in the images for the {\em dining, party} and {\em speech} classes vary significantly. Figure \ref{fig:dataset} illustrates example images from this dataset.

\subsection{Implementation Details}
For building classifiers, we use Support Vector Machine (SVM) classifiers with Gaussian-RBF kernel for each feature type. In order to form the combined facial descriptor, the proposed facial descriptors are simply concatenated together. We evaluate the recognition results in terms of average precision (AP) using five-fold cross validation over the whole dataset. We select a single set of parameters, including the kernel bandwidth and SVM cost parameter, using a greedy search over the training set, and use the same parameter set in all folds, for all classes.

Classifiers based on facial descriptors are combined with other feature classifiers via late fusion. For this, we learn a second layer linear SVM over the scores of the individual feature classifiers, and use its output as the final prediction.

\subsection{Results and Discussions}

In this section, we first present an evaluation of the proposed facial descriptors in detail. Then, we evaluate the combinations of the facial descriptors together with GIST, BoW on dense SIFT and CNN features.

\begin{table*}
\centering
\caption{Average Precisions of the individual face descriptors using the face detection outputs}
	\small{
	\begin{tabular}{ c|cccccccccc|c }	
		Feat. & b\&p & din. & h.sh. & h.fv. & hug. & kick. & kiss. & pty. & sp. & tlk. & AVG\\
		\hline
		HFO & 12.1 & 21.9 & 21.0 & 10.0 & 15.9 & 18.6 & 17.3 & 40.2 & 36.6 & 13.5  & 20.7 \\
		HFD & \textbf{15.7} & 20.6 & 21.5 & \textbf{15.7} &  8.3 & 22.7 & 18.2 & 42.7 & \textbf{37.8} & 16.8  & 22.0 \\
		DF & 15.6 & 24.6 & 23.3 & 12.1 & 14.3 &  5.9 & 23.7 & 23.4 & 21.8 & 13.2  & 17.8 \\
		CHFL & 14.9 & 19.3 & 19.9 & 17.3 & 13.3 & 13.3 & 13.9 & 17.7 & 16.3 &  8.9  & 15.5 \\
		GHFL & 14.8 & 25.2 & 22.3 & 12.4 & \textbf{22.8} & \textbf{22.8} & 18.2 & 27.0 & 18.1 & 14.1  & 19.8 \\
		\hline
		Comb. & 14.2 & \textbf{30.0} & \textbf{24.8} &  9.1 & 22.5 & 15.4 & \textbf{48.0} & \textbf{44.4} & 34.7 & \textbf{17.9}  & \textbf{26.1} \\
		\hline
	\end{tabular}
	}
\label{table:FaceDescAccuracies1}
\end{table*}

To evaluate the performance of facial descriptors, we have two settings: 1) Automatic: We use the output of the face detectors, 2) Face ground truth: We manually label each face region, together with the correct orientation of the face. We first present the experimental results using the automatic setting. The first five results in Table \ref{table:FaceDescAccuracies1} presents the recognition rate using the proposed facial descriptors individually, using the face detector outputs. According to these results, Histogram of Face Directions (HFD) gives the best performance, amongst the individual descriptors, whereas Circular Histogram of Face Locations (CHFL) performs the worst. Overall, the combination of all descriptors yields the best performance. We also observe that especially \textit{high five, boxing-punching} and \textit{kicking} classes are difficult to identify. In these interactions, the relative spatial distribution of the faces tend to have a more random structure, and this affects the classification performance.
 
\begin{table*}
\centering
\small{
\caption{Average Precisions of the individual face descriptors using ground truth faces}
	\begin{tabular}{ c|cccccccccc|c }
	
		Feat. & b\&p & din. & h.sh. & h.fv. & hug. & kick. & kiss. & pty. & sp. & tlk. & AVG\\
		\hline
		HFO & 14.0 & 40.8 & 31.7 & 13.0 & 49.6 &  9.3 & 24.0 & 54.9 & \textbf{91.8} & 14.5  & 34.4 \\
		HFD & 13.9 & 52.4 & 22.1 & \textbf{16.1} & 43.4 & 12.5 & 17.4 & 48.5 & 89.0 & 17.8  & 33.3 \\
		DF & 15.8 & 41.8 & 28.7 &  9.3 & 38.9 & \textbf{32.9} & 51.3 & 36.6 & 65.0 & \textbf{27.8}  & 34.8 \\
		CHFL & 12.8 & 33.2 & 20.5 & 13.1 & 25.4 & 12.0 & 13.8 & 38.8 & 72.6 & 16.0  & 25.8 \\
		GHFL & 10.7 & 48.1 & 21.9 &  6.2 & 32.4 & 17.4 &  8.2 & 46.0 & 70.1 & 14.2  & 27.5 \\
		\hline
		Comb. & \textbf{17.9} & \textbf{57.1} & \textbf{41.2} &  9.7 & \textbf{61.8} & 28.0 & \textbf{85.3} & \textbf{69.4} & 90.7 & 14.9  & \textbf{47.6} \\
		
			\hline
	\end{tabular}
	}
\label{table:FaceDescAccuraciesGt}
\end{table*}

By a qualitative evaluation of the dataset, we observe that the following conditions particularly affect the recognition performance: (1)~In photographs, people usually look at camera, so the natural orientations or directions of the faces can not be inferred. It is difficult to harvest images that depict natural occurrences of the interactions. (2)~Face detection algorithms may not find all the faces that are visible in the images. Although we use the state-of-the-art face detection algorithms, many false negatives occur. (3)~False positives also affect the performance of the facial descriptors negatively.

In order to evaluate the performance of the proposed descriptors independent of the errors introduced by the face detection, we manually annotated the faces in images with both location and orientation information, and extracted the ground-truth faces.  In this way, we aim to simulate the case with the perfect face detector. The results using this setting is presented in Table \ref{table:FaceDescAccuraciesGt}. The overall mean average precision of the combined descriptors with ground-truth faces rises from $26.1\%$ to $47.6 \%$. In other words, if we have perfect face detector and estimations for face orientations, we can achieve $47.6$ mAP, even with relatively simple 31-dimensional face descriptors. There are also some interesting observations; An average precision as high as $91.8\%$ is achievable using only HFO features for the \textit{speech} class. This is not surprising, as for the speech class, most of the faces are turned towards the speaker, and this situation forms a discriminative feature. In addition, for the \textit{kiss} interaction, the combination of the facial features achieves a quite high mAP of $85.3\%$.

Next, we evaluate the performance of the scene features for interaction recognition. We first evaluate the individual performances of GIST, BoW, BoW with spatial pyramid matching (SPM)~\cite{SPM}, and also the performances of deep features Places-CNN and Hybrid-CNN~\cite{CNN}. After that, we evaluate the results for combination of all feature sets and show the results in Table \ref{table:comparison1}. We also report the combinations of scene features with facial descriptors formed using the ground truth face detections in Table \ref{table:comparison2}.

\begin{table}
\caption{Comparison of the feature combinations using facial descriptors formed over the output of the face detectors.}
\centering
\begin{adjustbox}{max width=\linewidth}
\begin{tabular}{l|c|l}
\hline
		Type & Feature & AP \\
		\hline
		Regular & FaceDesc & 26.11 \\
		& GIST & 43.52 \\
		& BoW & 58.25 \\
		& BoW-SPM & 58.68 \\
		\hline
		Combined & BoW-SPM + GIST & 59.46 \\
		& FaceDesc + GIST & 47.13 \\
		& FaceDesc + BoW-SPM & 59.77 \\
		& FaceDesc + GIST + BoW-SPM & 60.60 \\
		\hline
		Deep & Places-CNN & 72.8 \\
		& Hybrid-CNN & 77.86 \\
		\hline 
		Deep Combined& FaceDesc + Hybrid-CNN & \textbf{78.33} \\
		& GIST + BoW-SPM + Hybrid-CNN & 75.05\\
		& FaceDesc + GIST + BoW-SPM + Hybrid-CNN & 76.09\\
		
		\hline
	\end{tabular}
\end{adjustbox}
\label{table:comparison1}
\end{table}

\begin{table}
\caption{Comparison of the feature combinations using facial descriptors formed over the ground truth faces.}
\centering 
\begin{adjustbox}{max width=\linewidth}
		\begin{tabular}{c|c}
		\hline
		Feature & AP \\
		\hline
		GtFaceDesc & 47.59 \\
		GtFaceDesc + GIST & 56.18 \\
		GtFaceDesc + BoW-SPM & 63.41 \\
		GtFaceDesc + GIST + BoW-SPM & 64.48 \\
		\hline
		GtFaceDesc + Hybrid-CNN &\textbf{80.11} \\
		GtFaceDesc + GIST + BoW-SPM + Hybrid-CNN & 79.56 \\
		\hline
	\end{tabular}
	\end{adjustbox}
\label{table:comparison2}
\end{table}

Results in Table \ref{table:comparison1} and Table~\ref{table:comparison2} show that, compared to the facial descriptors, scene features carry quite a lot of information for predicting the type of interaction. Proposed face descriptors contribute to the recognition of human interactions, and together with BoW and GIST features, they achieve a reasonable mAP of $60.6\%$, which would rise to $64.48\%$ in the presence of a perfect face detector. On the other hand, the deep features are the most effective for human interaction recognition, Places-CNN achieving $72.8\%$ AP and Hybrid-CNN feature achieving an AP of $77.86\%$. An interesting observation at this point is that, when combining deep features with regular scene descriptors such as GIST and BoW, the performance slightly drops. This result suggests that GIST or BoW features does not carry any complementary information to the deep features. On the contrary, combining proposed facial descriptors with CNN features improves the performance, achieving $78.33\%$ using output of face detectors, and $80.11\%$ when the ground truth face locations are used. 

These results show that, it is not easy to describe human interactions by looking at facial regions and their spatial layouts only (as demonstrated in Fig~\ref{fig:interactionsWithOnlyFaces}). Nevertheless, when used in combination with scene elements extracted from the whole image, the facial descriptors can boost the recognition performance.

\begin{figure}
\begin{center}
	\includegraphics[width=\linewidth]{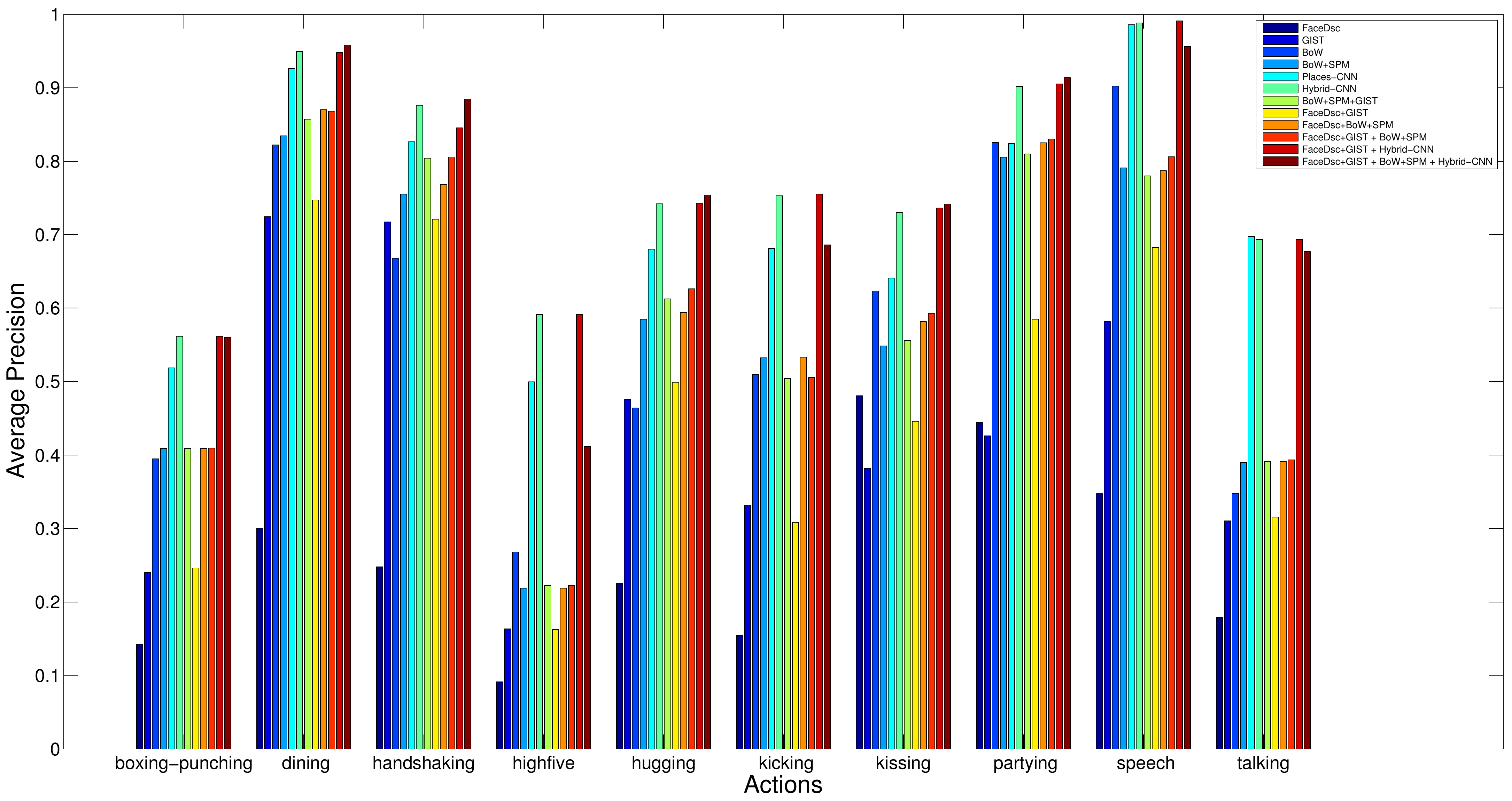}
\end{center}
\caption{Average precisions per each of the interaction classes. In calculation of facial descriptions \texttt{FaceDesc}, the outputs of the face detectors are used.}
\label{fig:APclassAll}
\end{figure}

\begin{figure}
\begin{center}
	\includegraphics[width=\linewidth]{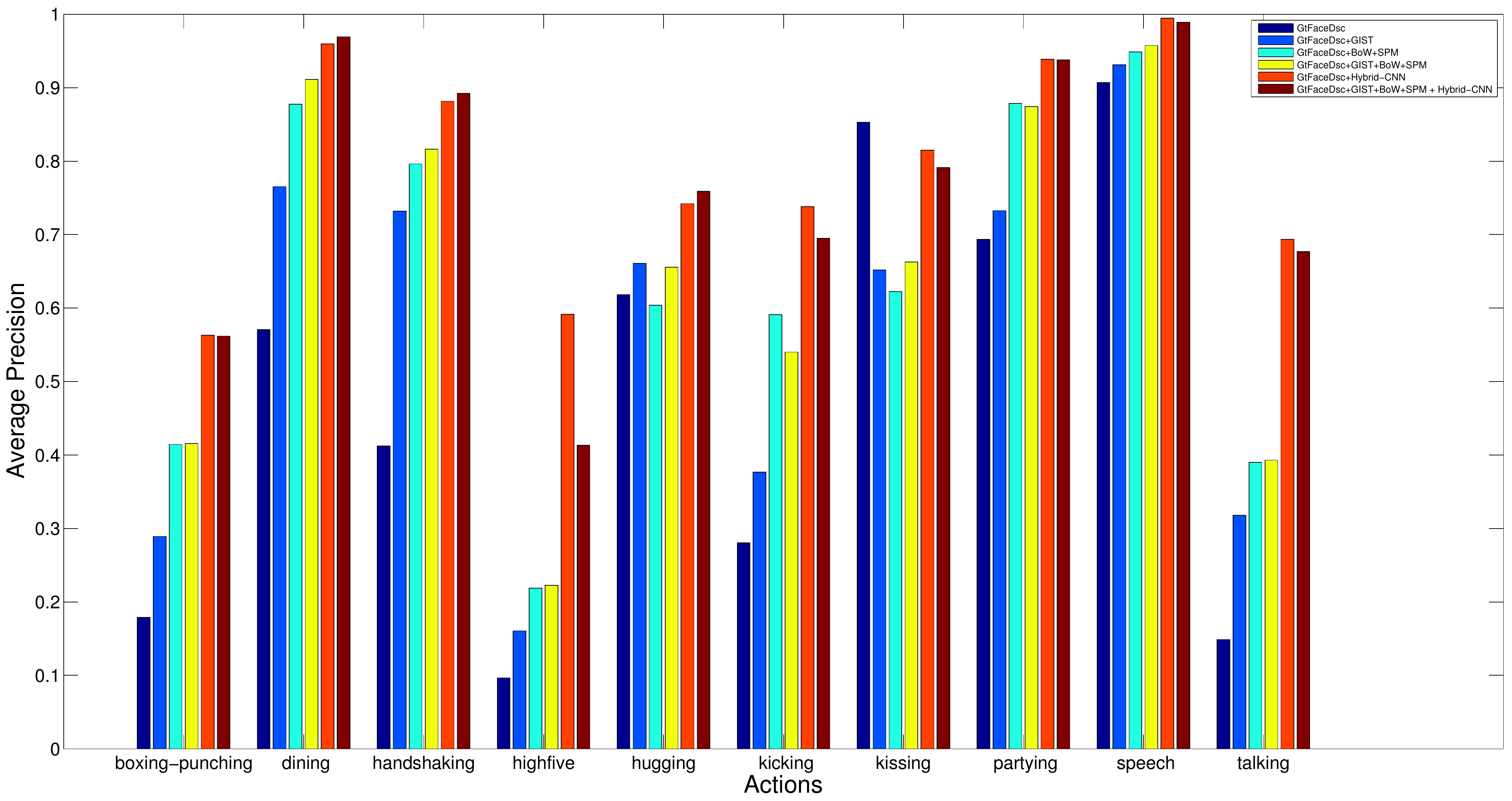}
\end{center}
\caption{Average precisions per each of the interaction classes. In calculation of facial descriptions \texttt{GtFaceDesc}, the ground truth face locations are used.}
\label{fig:APclassAllGT}
\end{figure}

We also present per class average precisions in Fig.~\ref{fig:APclassAll} and in Fig.~\ref{fig:APclassAllGT}. We observe that some of the interactions, such as boxing and high-five are more difficult to recognize, having average precisions lower than 0.6. On the other hand, recognition performance on some interaction classes, such as dining and speech, are quite high. This is likely to be due to the similar spatial configurations present in these interactions. In all of the interaction classes, CNN features are very effective.

\begin{figure*}
\begin{center}
\fboxsep=0mm
\fboxrule=1pt
    \fcolorbox{white}{white}{\includegraphics[width=0.085\linewidth,height=1.2cm]{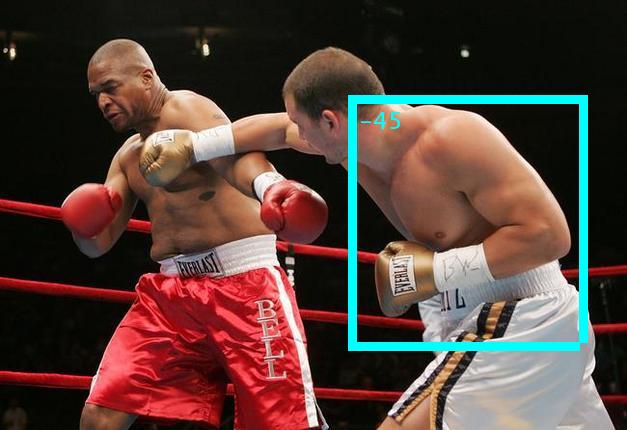}}
    \fcolorbox{white}{white}{\includegraphics[width=0.085\linewidth,height=1.2cm]{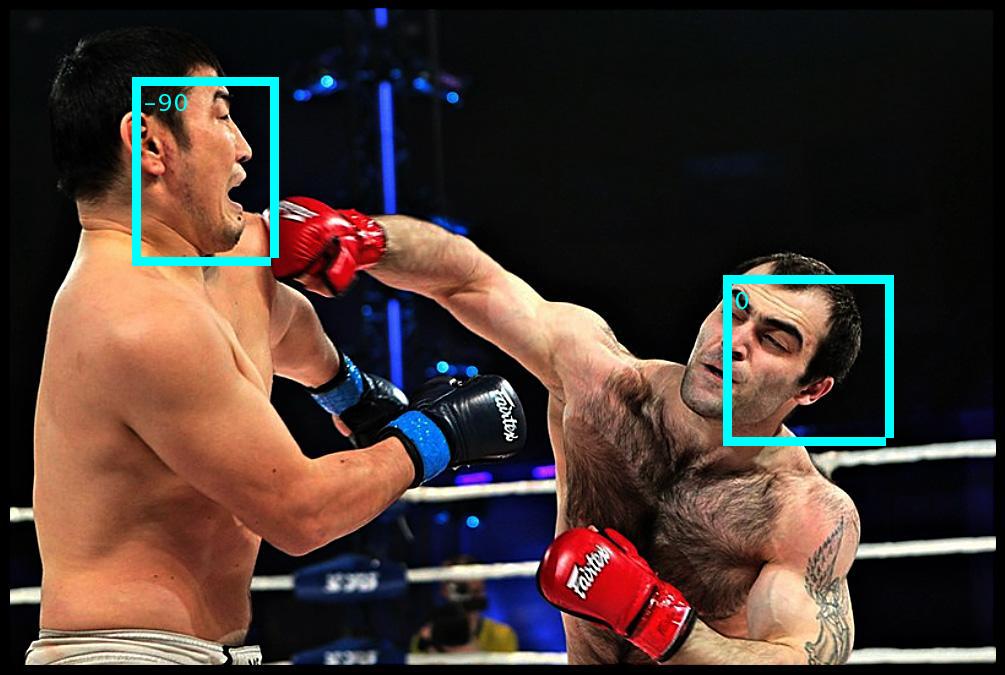}}
    \fcolorbox{white}{white}{\includegraphics[width=0.085\linewidth,height=1.2cm]{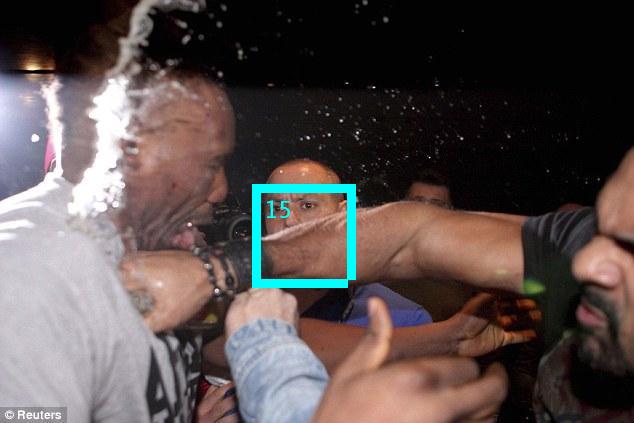}}
    \fcolorbox{red}{red}{\includegraphics[width=0.085\linewidth,height=1.2cm]{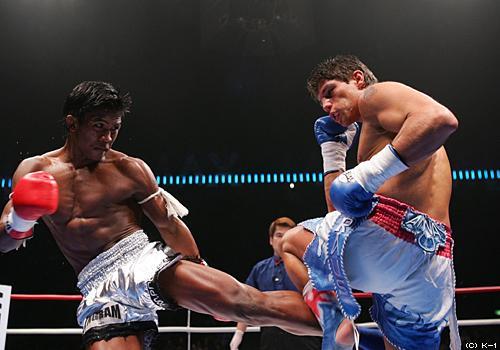}}
    \fcolorbox{white}{white}{\includegraphics[width=0.085\linewidth,height=1.2cm]{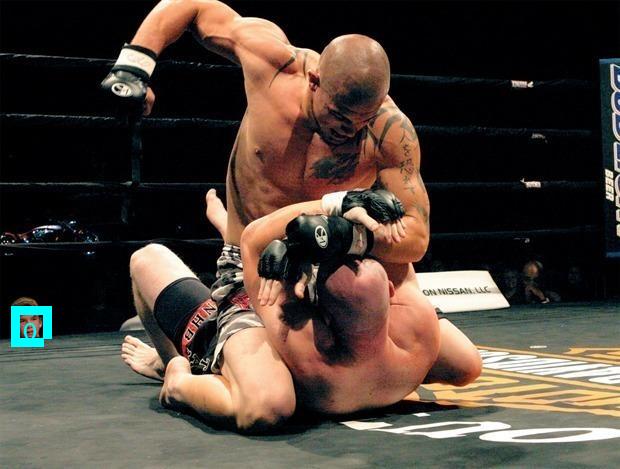}}
    \fcolorbox{white}{white}{\includegraphics[width=0.085\linewidth,height=1.2cm]{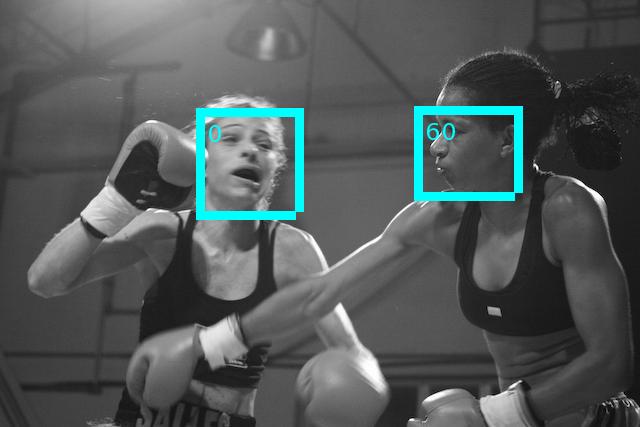}}
    \fcolorbox{white}{white}{\includegraphics[width=0.085\linewidth,height=1.2cm]{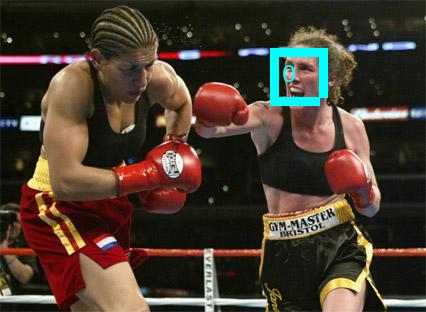}}
    \fcolorbox{white}{white}{\includegraphics[width=0.085\linewidth,height=1.2cm]{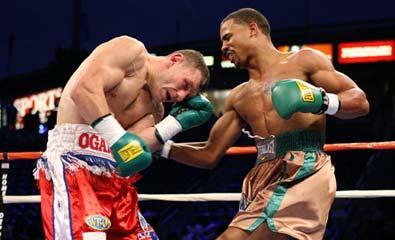}}
    \fcolorbox{red}{red}{\includegraphics[width=0.085\linewidth,height=1.2cm]{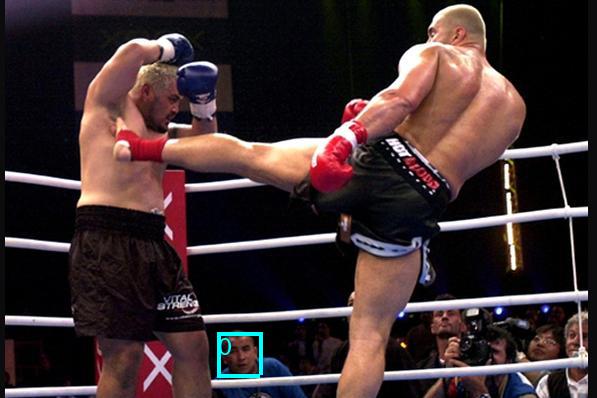}}
    \fcolorbox{white}{white}{\includegraphics[width=0.085\linewidth,height=1.2cm]{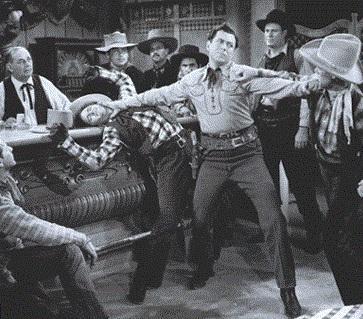}}\\
 
    \fcolorbox{white}{white}{\includegraphics[width=0.085\linewidth,height=1.2cm]{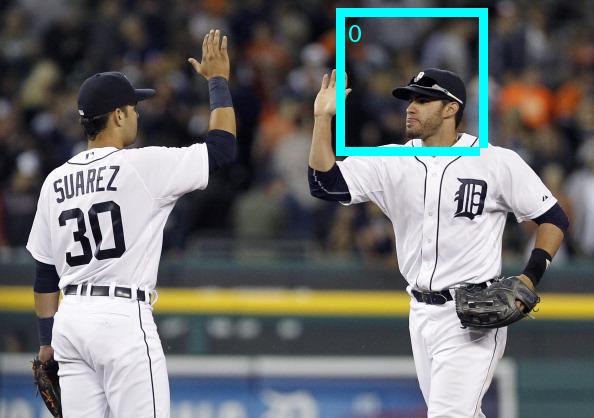}}
    \fcolorbox{red}{red}{\includegraphics[width=0.085\linewidth,height=1.2cm]{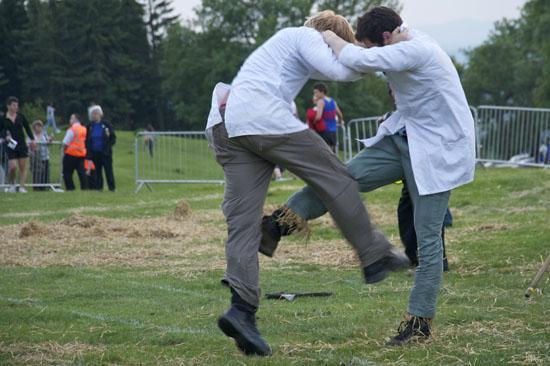}}
    \fcolorbox{red}{red}{\includegraphics[width=0.085\linewidth,height=1.2cm]{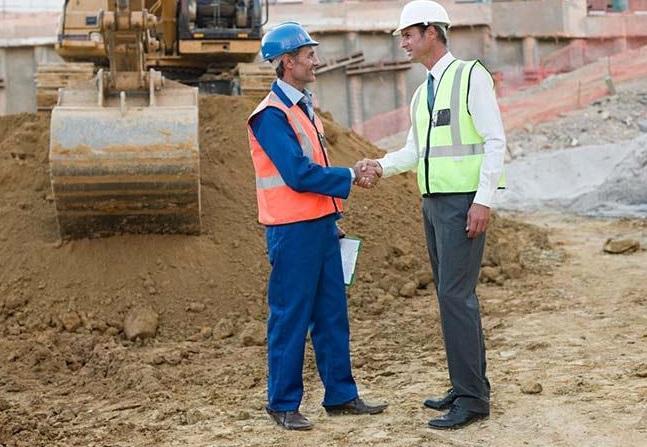}}
    \fcolorbox{red}{red}{\includegraphics[width=0.085\linewidth,height=1.2cm]{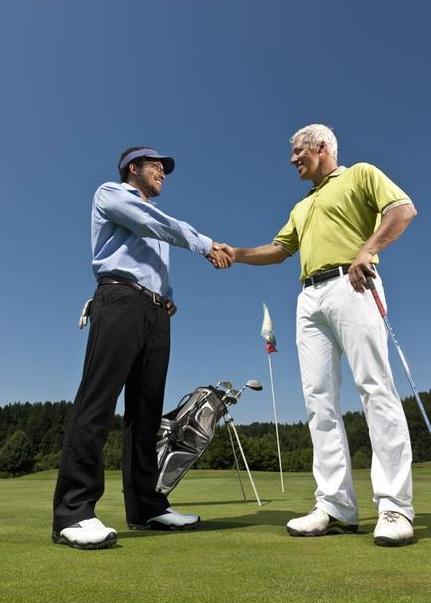}}
    \fcolorbox{white}{white}{\includegraphics[width=0.085\linewidth,height=1.2cm]{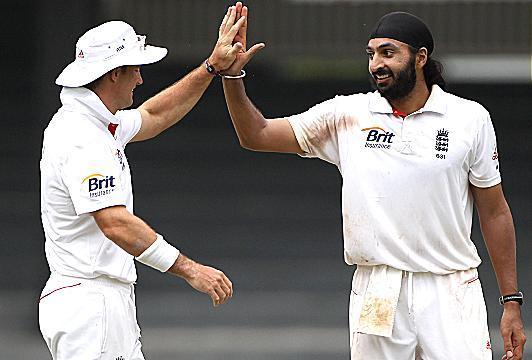}}
    \fcolorbox{red}{red}{\includegraphics[width=0.085\linewidth,height=1.2cm]{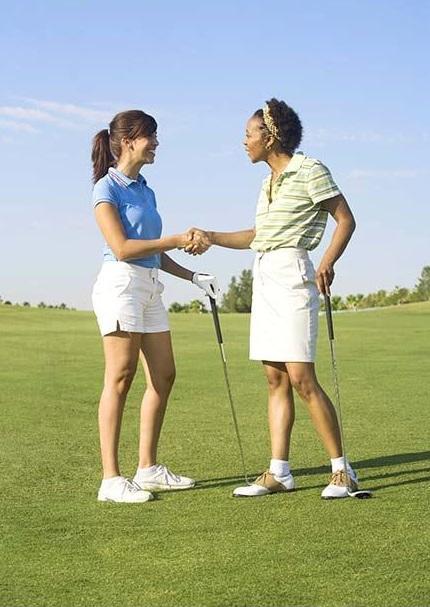}}
    \fcolorbox{white}{white}{\includegraphics[width=0.085\linewidth,height=1.2cm]{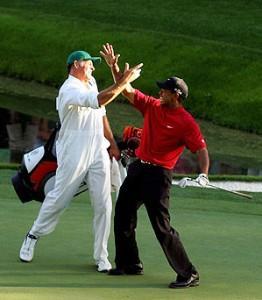}}
    \fcolorbox{white}{white}{\includegraphics[width=0.085\linewidth,height=1.2cm]{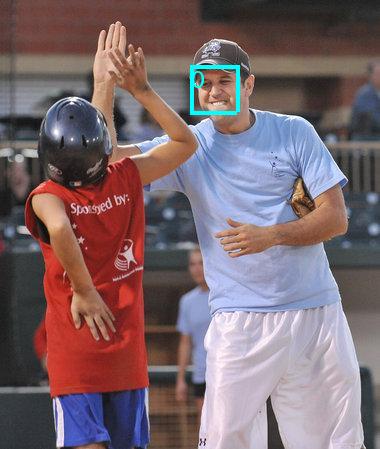}}
    \fcolorbox{white}{white}{\includegraphics[width=0.085\linewidth,height=1.2cm]{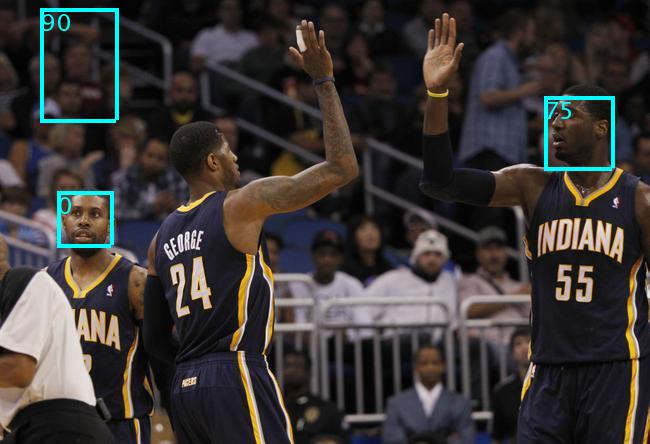}}
    \fcolorbox{red}{red}{\includegraphics[width=0.085\linewidth,height=1.2cm]{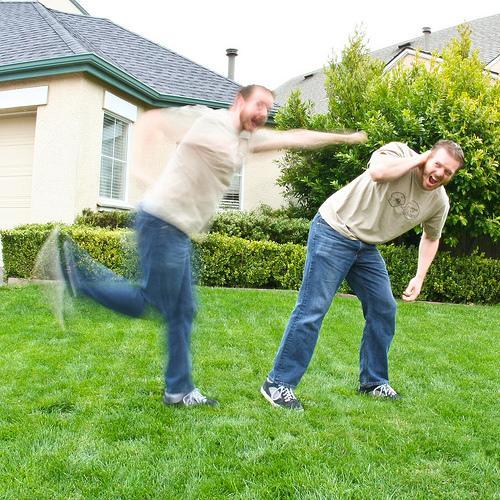}}\\
 
    \fcolorbox{white}{white}{\includegraphics[width=0.085\linewidth,height=1.2cm]{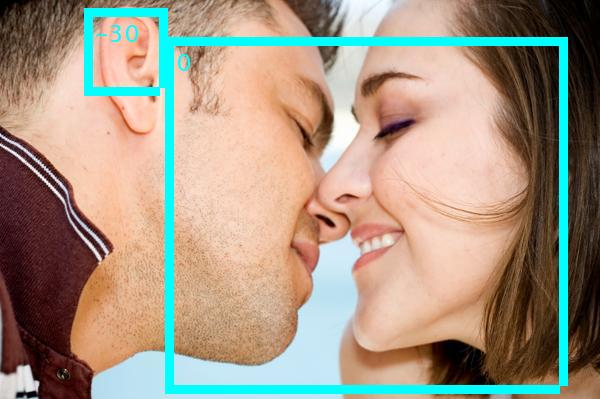}}
    \fcolorbox{white}{white}{\includegraphics[width=0.085\linewidth,height=1.2cm]{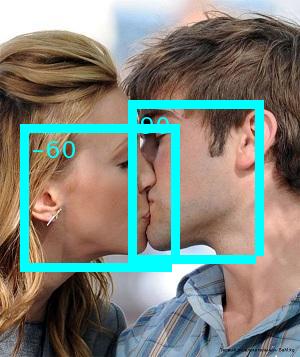}}
    \fcolorbox{white}{white}{\includegraphics[width=0.085\linewidth,height=1.2cm]{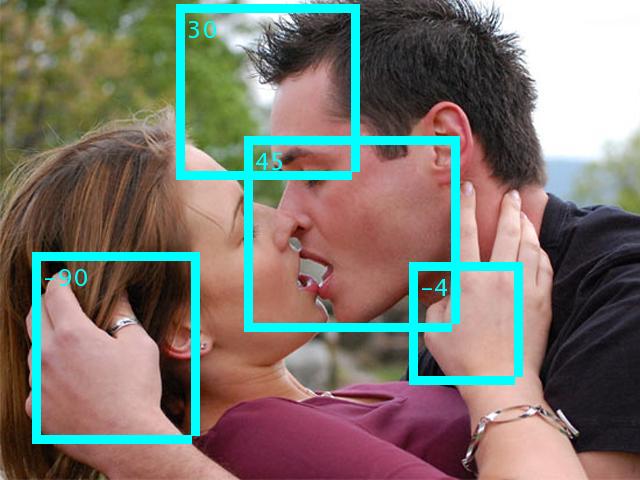}}
    \fcolorbox{white}{white}{\includegraphics[width=0.085\linewidth,height=1.2cm]{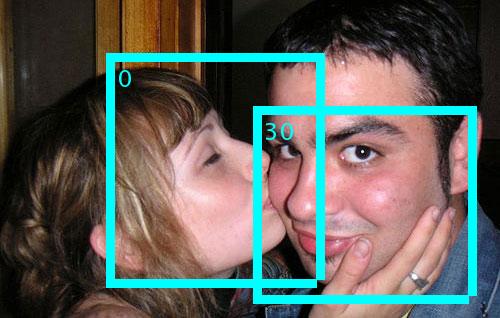}}
    \fcolorbox{white}{white}{\includegraphics[width=0.085\linewidth,height=1.2cm]{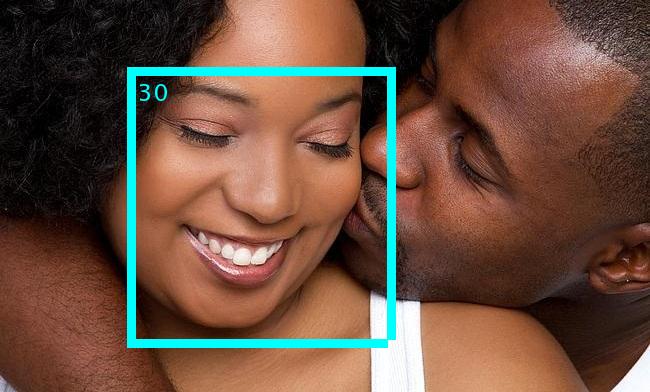}}
    \fcolorbox{white}{white}{\includegraphics[width=0.085\linewidth,height=1.2cm]{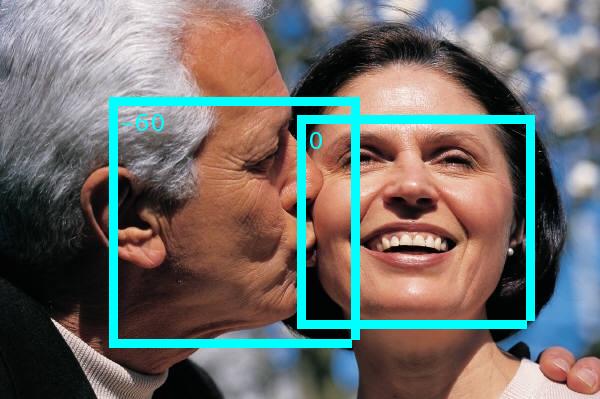}}
    \fcolorbox{white}{white}{\includegraphics[width=0.085\linewidth,height=1.2cm]{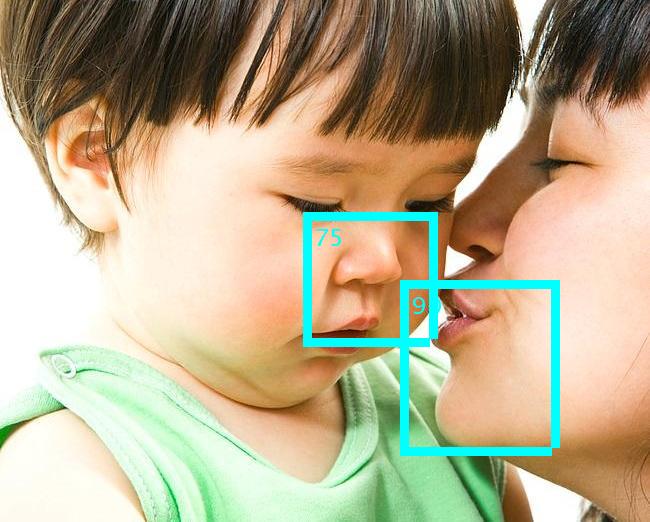}}
    \fcolorbox{white}{white}{\includegraphics[width=0.085\linewidth,height=1.2cm]{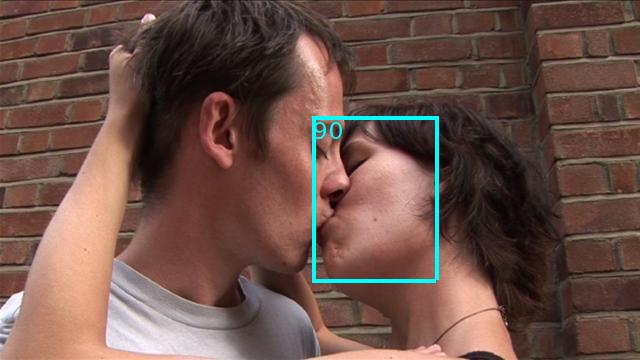}}
    \fcolorbox{white}{white}{\includegraphics[width=0.085\linewidth,height=1.2cm]{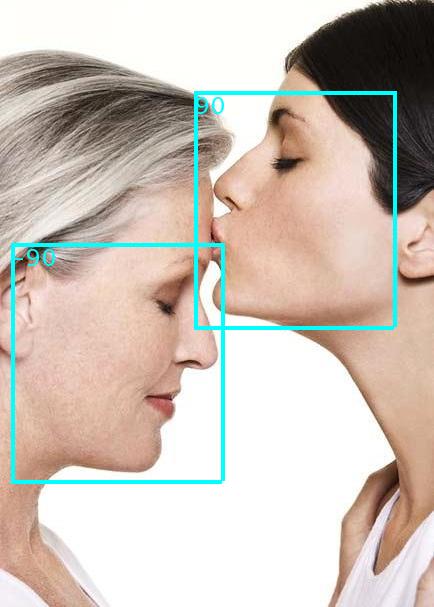}}
    \fcolorbox{red}{red}{\includegraphics[width=0.085\linewidth,height=1.2cm]{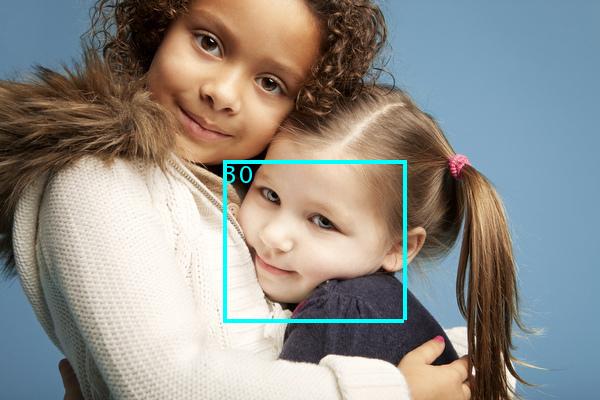}}\\
 
    \fcolorbox{white}{white}{\includegraphics[width=0.085\linewidth,height=1.2cm]{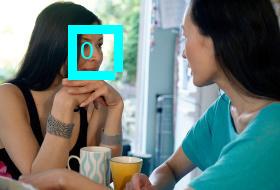}}
    \fcolorbox{white}{white}{\includegraphics[width=0.085\linewidth,height=1.2cm]{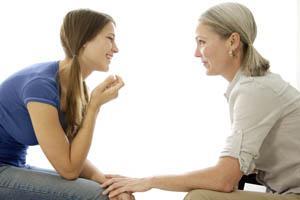}}
    \fcolorbox{white}{white}{\includegraphics[width=0.085\linewidth,height=1.2cm]{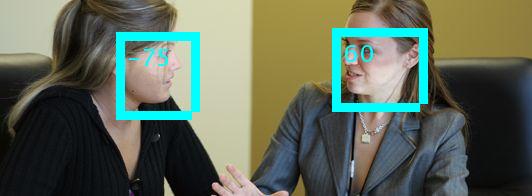}}
    \fcolorbox{white}{white}{\includegraphics[width=0.085\linewidth,height=1.2cm]{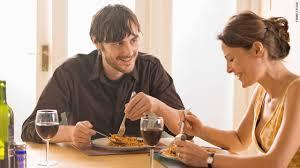}}
    \fcolorbox{white}{white}{\includegraphics[width=0.085\linewidth,height=1.2cm]{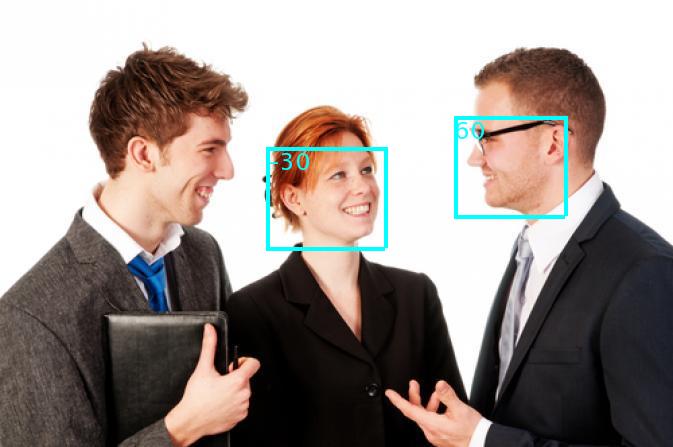}}
    \fcolorbox{white}{white}{\includegraphics[width=0.085\linewidth,height=1.2cm]{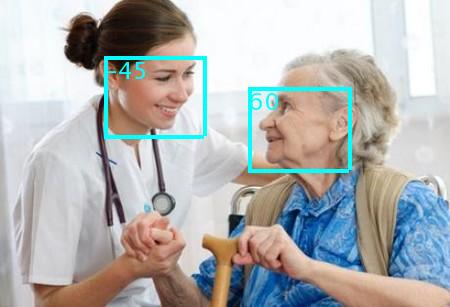}}
    \fcolorbox{white}{white}{\includegraphics[width=0.085\linewidth,height=1.2cm]{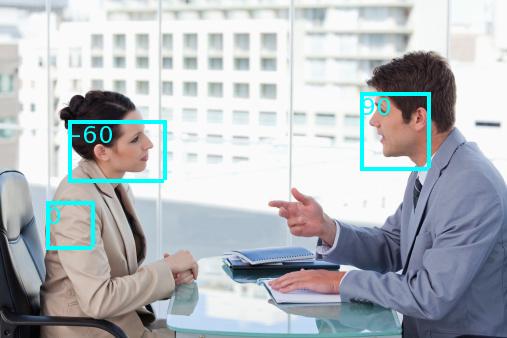}}
    \fcolorbox{white}{white}{\includegraphics[width=0.085\linewidth,height=1.2cm]{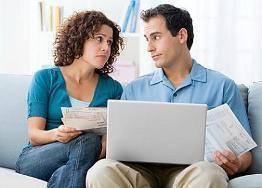}}
    \fcolorbox{white}{white}{\includegraphics[width=0.085\linewidth,height=1.2cm]{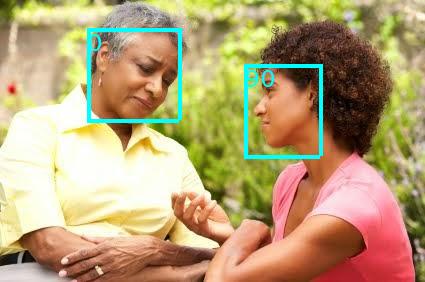}}
    \fcolorbox{white}{white}{\includegraphics[width=0.085\linewidth,height=1.2cm]{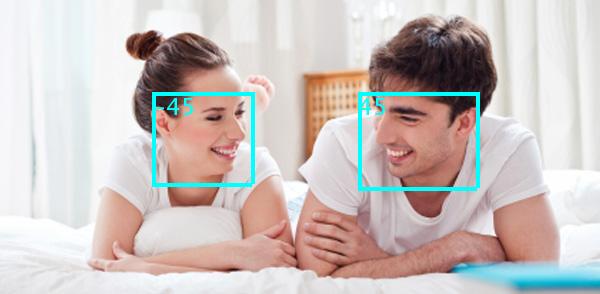}}\\
 
\end{center}
\vspace{-3mm}
	\caption{Images that have the top ten scores using \texttt{FaceDesc} extracted from automatic face detections, \texttt{GIST + BoW-SPM + Hybrid-CNN} features, for \textit{boxing-punching, high-five, kissing} and \textit{talking} interactions. Out-of-class images are shown with red borders and the face detection outputs (if any) are shown in cyan.}
\label{fig:qualEx}
\end{figure*}

Qualitative results that are obtained as a result of using \texttt{FaceDesc + GIST + BoW-SPM + Hybrid-CNN} combination are presented in Fig.~\ref{fig:qualEx}. As it can be seen in these images, high five interaction is mostly confused with handshaking and punching, whereas kissing interaction is mostly confused with hugging. 

One of the important characteristics of the proposed facial descriptors is their efficiency. Table~\ref{table:DescDimensions} shows the dimensionalities of each of the utilized descriptors. Our proposed 31-dimensional face descriptors provides 2-6\% points increase in average precision when used with BoW+SPM and/or GIST, and 1-3\% points increase when used with the CNN features, without requiring excessive training.

\begin{table}
\centering
\caption{Feature dimensions of the utilized descriptors}
\small{
	\begin{tabular}{ c|ccccc }	
		Feature & FaceDesc & GIST & BoW & BoW+SPM  &CNN \\
		\hline
		\# of Dimensions & 31 & 512 & 1000 & 4000  & 4096 \\
		\hline
		\end{tabular}
		}
\label{table:DescDimensions}
\end{table}


\section{Conclusion}
 In this paper, we look into a rarely studied area of computer vision, namely human interaction recognition in still images. We investigate whether we can infer the correct label of an interaction image by looking at the facial regions, their relative positions and spatial layout. In order to capture such information, we propose several descriptors based on facial regions. Our experimental results show that, facial descriptors provide meaningful information, however, using them in isolation yields less effective results. When combined with global scene features, especially deep features, proposed facial descriptors are shown to have improved recognition performance. 
 
In this context, we introduce a new image dataset which can be used for human interaction recognition, and also for evaluating face detectors performance on further tasks. The faces in the dataset are annotated with both locations and orientations, and will be made available upon publication.

\section*{Acknowledgements}
This work was supported in part by the Scientific and Technological Research Council of Turkey (TUBITAK) Career Development Award numbered 112E149.


\bibliographystyle{model2-names}

\begin{thebibliography}{26}
\expandafter\ifx\csname natexlab\endcsname\relax\def\natexlab#1{#1}\fi
\providecommand{\url}[1]{\texttt{#1}}
\providecommand{\href}[2]{#2}
\providecommand{\path}[1]{#1}
\providecommand{\DOIprefix}{doi:}
\providecommand{\ArXivprefix}{arXiv:}
\providecommand{\URLprefix}{URL: }
\providecommand{\Pubmedprefix}{pmid:}
\providecommand{\doi}[1]{\href{http://dx.doi.org/#1}{\path{#1}}}
\providecommand{\Pubmed}[1]{\href{pmid:#1}{\path{#1}}}
\providecommand{\bibinfo}[2]{#2}
\ifx\xfnm\relax \def\xfnm[#1]{\unskip,\space#1}\fi
\bibitem[{Aggarwal and Ryoo(2011)}]{Aggarwal2011}
\bibinfo{author}{Aggarwal, J.}, \bibinfo{author}{Ryoo, M.},
  \bibinfo{year}{2011}.
\newblock \bibinfo{title}{Human activity analysis: A review}.
\newblock \bibinfo{journal}{ACM Comput. Surv.} \bibinfo{volume}{43}.
\bibitem[{Bossard et~al.(2013)Bossard, Guillaumin and Van~Gool}]{bossard13}
\bibinfo{author}{Bossard, L.}, \bibinfo{author}{Guillaumin, M.},
  \bibinfo{author}{Van~Gool, L.}, \bibinfo{year}{2013}.
\newblock \bibinfo{title}{Event recognition in photo collections with a
  stopwatch hmm}, in: \bibinfo{booktitle}{IEEE International Conference on
  Computer Vision}.
\bibitem[{Delaitre et~al.(2010)Delaitre, Laptev and Sivic}]{Delaitre10}
\bibinfo{author}{Delaitre, V.}, \bibinfo{author}{Laptev, I.},
  \bibinfo{author}{Sivic, J.}, \bibinfo{year}{2010}.
\newblock \bibinfo{title}{Recognizing human actions in still images: a study of
  bag-of-features and part-based representations}, in:
  \bibinfo{booktitle}{BMVC}.
\bibitem[{Delaitre et~al.(2011)Delaitre, Sivic and Laptev}]{Delaitre2011}
\bibinfo{author}{Delaitre, V.}, \bibinfo{author}{Sivic, J.},
  \bibinfo{author}{Laptev, I.}, \bibinfo{year}{2011}.
\newblock \bibinfo{title}{Learning person-object interactions for action
  recognition in still images}, in: \bibinfo{booktitle}{Proc. NIPS}.
\bibitem[{Desai et~al.(2010)Desai, Ramanan and Fowlkes}]{desai2010}
\bibinfo{author}{Desai, C.}, \bibinfo{author}{Ramanan, D.},
  \bibinfo{author}{Fowlkes, C.}, \bibinfo{year}{2010}.
\newblock \bibinfo{title}{Discriminative models for static human-object
  interactions}, in: \bibinfo{booktitle}{Workshop on Structured Models in
  Computer Vision}.
\bibitem[{Fathi et~al.(2012)Fathi, Hodgins and Rehg}]{6247805}
\bibinfo{author}{Fathi, A.}, \bibinfo{author}{Hodgins, J.},
  \bibinfo{author}{Rehg, J.}, \bibinfo{year}{2012}.
\newblock \bibinfo{title}{Social interactions: A first-person perspective}, in:
  \bibinfo{booktitle}{{IEEE} Conference on Computer Vision and Pattern
  Recognition}, \bibinfo{publisher}{IEEE}, \bibinfo{address}{Los Alamitos, CA,
  USA}. pp. \bibinfo{pages}{1226--1233}.
\bibitem[{Gupta et~al.(2009)Gupta, Kembhavi and Davis}]{Gupta2009}
\bibinfo{author}{Gupta, A.}, \bibinfo{author}{Kembhavi, A.},
  \bibinfo{author}{Davis, L.}, \bibinfo{year}{2009}.
\newblock \bibinfo{title}{Observing human-object interactions: Using spatial
  and functional compatibility for recognition}.
\newblock \bibinfo{journal}{Pattern Analysis and Machine Intelligence, IEEE
  Transactions on} \bibinfo{volume}{31}, \bibinfo{pages}{1775--1789}.
\bibitem[{Hoai and Zisserman(2014)}]{HoaiCVPR14}
\bibinfo{author}{Hoai, M.}, \bibinfo{author}{Zisserman, A.},
  \bibinfo{year}{2014}.
\newblock \bibinfo{title}{Talking heads: Detecting humans and recognizing their
  interactions}, in: \bibinfo{booktitle}{Proceedings of IEEE Conference on
  Computer Vision and Pattern Recognition}.
\bibitem[{Jain et~al.(2013)Jain, Jegou and Bouthemy}]{betterMotion}
\bibinfo{author}{Jain, M.}, \bibinfo{author}{Jegou, H.},
  \bibinfo{author}{Bouthemy, P.}, \bibinfo{year}{2013}.
\newblock \bibinfo{title}{Better exploiting motion for better action
  recognition.}, in: \bibinfo{booktitle}{CVPR}, \bibinfo{publisher}{IEEE}. pp.
  \bibinfo{pages}{2555--2562}.
\bibitem[{Lazebnik et~al.(2006)Lazebnik, Schmid and Ponce}]{SPM}
\bibinfo{author}{Lazebnik, S.}, \bibinfo{author}{Schmid, C.},
  \bibinfo{author}{Ponce, J.}, \bibinfo{year}{2006}.
\newblock \bibinfo{title}{Beyond bags of features: Spatial pyramid matching for
  recognizing natural scene categories}, in: \bibinfo{booktitle}{Computer
  Vision and Pattern Recognition, 2006 IEEE Computer Society Conference on},
  pp. \bibinfo{pages}{2169--2178}.
\bibitem[{Li et~al.(2013)Li, Yu, Divakaran and
  Vasconcelos}]{Li.dynpooling.2013}
\bibinfo{author}{Li, W.}, \bibinfo{author}{Yu, Q.}, \bibinfo{author}{Divakaran,
  A.}, \bibinfo{author}{Vasconcelos, N.}, \bibinfo{year}{2013}.
\newblock \bibinfo{title}{Dynamic pooling for complex event recognition}, in:
  \bibinfo{booktitle}{Proceedings of IEEE International Conference on Computer
  Vision}, pp. \bibinfo{pages}{2728--2735}.
\bibitem[{Marin-Jimenez et~al.(2014)Marin-Jimenez, Zisserman, Eichner and
  Ferrari}]{JimenezIJCV2014}
\bibinfo{author}{Marin-Jimenez, M.}, \bibinfo{author}{Zisserman, A.},
  \bibinfo{author}{Eichner, M.}, \bibinfo{author}{Ferrari, V.},
  \bibinfo{year}{2014}.
\newblock \bibinfo{title}{Detecting people looking at each other in videos}.
\newblock \bibinfo{journal}{International Journal of Computer Vision}
  \bibinfo{volume}{106}, \bibinfo{pages}{282--296}.
\bibitem[{Oliva and Torralba(2001)}]{gist2001}
\bibinfo{author}{Oliva, A.}, \bibinfo{author}{Torralba, A.},
  \bibinfo{year}{2001}.
\newblock \bibinfo{title}{Modeling the shape of the scene: A holistic
  representation of the spatial envelope}.
\newblock \bibinfo{journal}{International Journal of Computer Vision}
  \bibinfo{volume}{42}, \bibinfo{pages}{145--175}.
\bibitem[{Park and Aggarwal(2000)}]{905274}
\bibinfo{author}{Park, S.}, \bibinfo{author}{Aggarwal, J.},
  \bibinfo{year}{2000}.
\newblock \bibinfo{title}{Recognition of human interaction using multiple
  features in gray scale images}, in: \bibinfo{booktitle}{Pattern Recognition,
  Proceedings. 15th International Conference on}, pp. \bibinfo{pages}{51 --54
  vol.1}.
\bibitem[{Park and Aggarwal(2006)}]{ParkAggar2006}
\bibinfo{author}{Park, S.}, \bibinfo{author}{Aggarwal, J.K.},
  \bibinfo{year}{2006}.
\newblock \bibinfo{title}{Simultaneous tracking of multiple body parts of
  interacting persons}.
\newblock \bibinfo{journal}{Comput. Vis. Image Underst.} \bibinfo{volume}{102},
  \bibinfo{pages}{1--21}.
\bibitem[{Prest et~al.(2012)Prest, Schmid and Ferrari}]{prest2012}
\bibinfo{author}{Prest, A.}, \bibinfo{author}{Schmid, C.},
  \bibinfo{author}{Ferrari, V.}, \bibinfo{year}{2012}.
\newblock \bibinfo{title}{Weakly supervised learning of interactions between
  humans and objects}.
\newblock \bibinfo{journal}{IEEE TPAMI} \bibinfo{volume}{34},
  \bibinfo{pages}{601--614}.
\bibitem[{Ramanathan et~al.(2013)Ramanathan, Yao and Fei-Fei}]{socialRole}
\bibinfo{author}{Ramanathan, V.}, \bibinfo{author}{Yao, B.},
  \bibinfo{author}{Fei-Fei, L.}, \bibinfo{year}{2013}.
\newblock \bibinfo{title}{Social role discovery in human events}, in:
  \bibinfo{booktitle}{Proceedings of the 2013 IEEE Conference on Computer
  Vision and Pattern Recognition}, pp. \bibinfo{pages}{2475--2482}.
\bibitem[{Ryoo and Aggarwal(2009)}]{RyooAggar2009}
\bibinfo{author}{Ryoo, M.S.}, \bibinfo{author}{Aggarwal, J.},
  \bibinfo{year}{2009}.
\newblock \bibinfo{title}{Spatio-temporal relationship match: Video structure
  comparison for recognition of complex human activities}, in:
  \bibinfo{booktitle}{Computer Vision, 2009 IEEE 12th International Conference
  on}, pp. \bibinfo{pages}{1593--1600}.
\bibitem[{Stottinger et~al.(2012)Stottinger, Uijlings, Pandey, Sebe and
  Giunchiglia}]{6248037}
\bibinfo{author}{Stottinger, J.}, \bibinfo{author}{Uijlings, J.},
  \bibinfo{author}{Pandey, A.}, \bibinfo{author}{Sebe, N.},
  \bibinfo{author}{Giunchiglia, F.}, \bibinfo{year}{2012}.
\newblock \bibinfo{title}{(unseen) event recognition via semantic
  compositionality}, in: \bibinfo{booktitle}{Computer Vision and Pattern
  Recognition (CVPR)}, pp. \bibinfo{pages}{3061--3068}.
\bibitem[{Viola and Jones(2004)}]{Viola2004}
\bibinfo{author}{Viola, P.}, \bibinfo{author}{Jones, M.J.},
  \bibinfo{year}{2004}.
\newblock \bibinfo{title}{Robust real-time face detection}.
\newblock \bibinfo{journal}{Int. J. Comput. Vision} \bibinfo{volume}{57},
  \bibinfo{pages}{137--154}.
\bibitem[{Wang and Schmid(2013)}]{wang2013}
\bibinfo{author}{Wang, H.}, \bibinfo{author}{Schmid, C.}, \bibinfo{year}{2013}.
\newblock \bibinfo{title}{{Action Recognition with Improved Trajectories}}, in:
  \bibinfo{booktitle}{{ICCV 2013 - IEEE International Conference on Computer
  Vision}}, \bibinfo{publisher}{IEEE}. pp. \bibinfo{pages}{3551--3558}.
\bibitem[{Yang et~al.(2012)Yang, Baker, Kannan and Ramanan}]{YangBKR12}
\bibinfo{author}{Yang, Y.}, \bibinfo{author}{Baker, S.},
  \bibinfo{author}{Kannan, A.}, \bibinfo{author}{Ramanan, D.},
  \bibinfo{year}{2012}.
\newblock \bibinfo{title}{Recognizing proxemics in personal photos.}, in:
  \bibinfo{booktitle}{CVPR}, \bibinfo{publisher}{IEEE}. pp.
  \bibinfo{pages}{3522--3529}.
\bibitem[{Yao and Fei-Fei(2010a)}]{YaoFeiFeiCVPR2010a}
\bibinfo{author}{Yao, B.}, \bibinfo{author}{Fei-Fei, L.},
  \bibinfo{year}{2010}a.
\newblock \bibinfo{title}{Grouplet: a structured image representation for
  recognizing human and object interactions}, in: \bibinfo{booktitle}{The
  Twenty-Third IEEE Conference on Computer Vision and Pattern Recognition},
  \bibinfo{address}{San Francisco, CA}.
\bibitem[{Yao and Fei-Fei(2010b)}]{YaoFeiFeiCVPR2010b}
\bibinfo{author}{Yao, B.}, \bibinfo{author}{Fei-Fei, L.},
  \bibinfo{year}{2010}b.
\newblock \bibinfo{title}{Modeling mutual context of object and human pose in
  human-object interaction activities}, in: \bibinfo{booktitle}{CVPR},
  \bibinfo{address}{San Francisco, CA}.
\bibitem[{Zhou et~al.(2014)Zhou, Lapedriza, Xiao, Torralba and Oliva}]{CNN}
\bibinfo{author}{Zhou, B.}, \bibinfo{author}{Lapedriza, A.},
  \bibinfo{author}{Xiao, J.}, \bibinfo{author}{Torralba, A.},
  \bibinfo{author}{Oliva, A.}, \bibinfo{year}{2014}.
\newblock \bibinfo{title}{Learning deep features for scene recognition using
  places database}, in: \bibinfo{booktitle}{NIPS}.
\bibitem[{Zhu and Ramanan(2012)}]{FaceDetection1}
\bibinfo{author}{Zhu, X.}, \bibinfo{author}{Ramanan, D.}, \bibinfo{year}{2012}.
\newblock \bibinfo{title}{Face detection, pose estimation and landmark
  localization in the wild}.
\newblock \bibinfo{note}{CVPR}.

\end{thebibliography}

\end{document}